\documentclass[letterpaper, 10 pt, conference]{ieeeconf}  % Comment this line out if you need a4paper

\IEEEoverridecommandlockouts                              % This command is only needed if 
\usepackage[backend=biber,style=ieee,sorting=none,giveninits=true,maxbibnames=99,doi=false,
url=false,eprint=false,isbn=false]{biblatex}

\AtBeginBibliography{\small}

\usepackage[T1]{fontenc}
\usepackage[utf8]{inputenc}
\usepackage{amsmath,amssymb}
\usepackage{tabularray}
\usepackage{graphicx}
\usepackage{float}
\usepackage{dblfloatfix}
\title{\LARGE \bf
Battery-Aware Predictive Trajectory Planning and Control for Multirotors Under Disturbances}

\author{Krishna Bhavithavya Kidambi$^{1}$ % <-this % stops a space
\thanks{$^{1}$Kidambi, K.B is with department of Mechanical and Aerospace Engineering at University of Dayton, Dayton, OH. {\tt\small kkidambi1@udayton.edu}}%
}

\begin{document}

\maketitle
\thispagestyle{empty}
\pagestyle{empty}

%%%%%%%%%%%%%%%%%%%%%%%%%%%%%%%%%%%%%%%%%%%%%%%%%%%%%%%%%%%%%%%%%%%%%%%%%%%%%%%%
\begin{abstract}
	This paper presents a battery-aware predictive trajectory-planning and control framework
	for multirotors operating under spatially localized disturbances.
	Candidate trajectories are evaluated through closed-loop
	vehicle--motor--battery propagation, allowing disturbance-induced control
	demand, electrical energy, battery evolution, and terminal-voltage-dependent
	actuator capability to enter the planning process.
	A reduced-order battery model is numerically benchmarked against an
	independently implemented Simscape equivalent-circuit reference, with a power
	NRMSE of $0.64\%$ and a cumulative-energy discrepancy below $0.7\%$.
	In a $150$-s, $640$-m mission containing three disturbance regions, the
	selected trajectory reduces electrical energy consumption by $7.46\%$ and
	position-tracking RMSE by approximately $72\%$ relative to the
	disturbance-aware fixed-reference baseline.
	Planner ablations show that battery-dependent terms are nonbinding at nominal
	SOC but alter the selected trajectory under a depleted-battery stress
	condition.
	Execution with multiple feedback controllers further demonstrates that
	controller selection changes the tradeoff among tracking accuracy, energy
	consumption, and actuator utilization.
	The results demonstrate the benefit of accounting for predicted closed-loop
	energetic and battery--actuator consequences during trajectory selection.
\end{abstract}
%%%%%%%%%%%%%%%%%%%%%%%%%%%%%%%%%%%%%%%%%%%%%%%%%%%%%%%%%%%%%%%%%%%%%%%%%%%%%%%%
\section{INTRODUCTION} 
Multirotor unmanned aerial vehicles (UAVs) are increasingly expected to
perform sustained autonomous missions in uncertain and disturbance-prone
environments \cite{idrissiReviewQuadrotorUnmanned2022,
	senguptaUrbanAirMobility2025}.
For battery-powered multirotors, however, mission performance is fundamentally
constrained by the finite onboard energy supply and the propulsion demand
required to execute the prescribed trajectory.
Environmental disturbances can substantially increase feedback-control effort
and rotor demand, accelerating battery discharge and reducing the terminal
voltage available to the propulsion system.
Consequently, as a mission progresses, the trajectory, disturbance exposure,
closed-loop control demand, battery state, and available actuator authority
become increasingly coupled.
Accounting for this interaction during trajectory planning is therefore
important for maintaining both energetic and closed-loop feasibility during
demanding multirotor missions.

Reliable multirotor operation depends not only on the feedback controller
used to regulate the vehicle, but also on the trajectory supplied to the
closed-loop system.
A wide range of linear and nonlinear control strategies have been developed
for multirotor trajectory tracking, including PID and linear-quadratic
methods, feedback linearization \cite{martinsInnerouterFeedbackLinearization2022}, sliding-mode control
\cite{choiAdaptiveNeuroFuzzySliding2025}, adaptive control
\cite{liuAdaptivePredefinedTimePrescribed2026}, and robust nonlinear control
\cite{johnstonAdaptiveModifiedRISE2025a}.
These methods have improved disturbance rejection and robustness to uncertain
vehicle dynamics, enabling increasingly aggressive flight
\cite{liuQuadrotorAggressiveControl2025}.
More recently, learning-based approaches have been investigated to estimate
unmodeled dynamics and disturbances online and adapt the closed-loop response
to operating conditions that are difficult to characterize a priori
\cite{guLearningUncertaintiesOnline2025,renLearningAgileQuadrotor2026}.
Despite these advances, feedback control is fundamentally tasked with realizing a prescribed reference. Consequently, its performance and required control effort remain strongly influenced by the trajectory supplied to the closed-loop system. 

Trajectory planning provides a complementary mechanism for improving
multirotor performance by determining how the vehicle interacts with its
environment before the resulting reference is executed by the feedback
controller.
Recent approaches have moved beyond conventional collision-free trajectory
generation to account explicitly for dynamic feasibility, computational
efficiency, sensing requirements, actuator limitations, and environmental
uncertainty.
Optimization-based methods have exploited the full multirotor dynamics to
generate time-optimal trajectories for aggressive flight, while probabilistic
inference methods have been developed to generate smooth and safe trajectories
under limited onboard computational resources
\cite{zhouEfficientRobustTimeOptimal2023,
	xingProbabilisticInferenceBasedEfficient2025}.
Physical and sensing constraints have also been incorporated directly into
the planning problem, including motor saturation and actuator degradation
\cite{zhangSafetyPlanningControl2023}, as well as localization uncertainty
and perception quality \cite{sunSafetyDrivenLocalizationUncertaintyDriven2024}.
Environmental disturbances have similarly been considered during trajectory
generation. Stochastic wind information has been incorporated into trajectory
optimization to reduce thrust-related costs
\cite{greiffQuadrotorMotionPlanning2023}, while learned disturbance models
have been embedded within predictive control to adapt motion plans to
uncertain wind and interaction-induced effects
\cite{lapandicMetaLearningAugmentedMPC2024}.
Collectively, these approaches demonstrate that trajectory planning can
proactively account for dynamic, actuator, sensing, and environmental
constraints rather than leaving their consequences entirely to the feedback
controller during execution.
However, these formulations generally do not explicitly propagate the complete closed-loop coupling by which trajectory-dependent disturbance exposure alters feedback-control and rotor demand, the resulting electrical load affects battery terminal voltage, and the evolving terminal voltage subsequently changes the actuator authority available during candidate-trajectory evaluation.

The finite onboard energy available to battery-powered multirotors introduces
a planning constraint that cannot be characterized solely by flight time or
geometric path length
\cite{gascheEnergyAwareSafe2025,yacefOptimizationEnergyConsumption2017}.
The electrical power required for flight depends on the vehicle operating
condition, including thrust, velocity, acceleration, payload, and
environmental effects such as wind
\cite{benarfaMotionEnergyHealthAware2026a}.
Accordingly, physics-based and data-driven approaches have been developed to
estimate multirotor power consumption and predict battery discharge during
flight
\cite{conteDatadrivenLearningMethod2022,daiDataefficientModelingPower2024}.
More detailed energy-aware formulations have coupled multirotor dynamics with
lithium-ion battery and powertrain models, allowing quantities such as state
of charge (SOC), terminal voltage, current, and polarization effects to evolve
with propulsion demand
\cite{baekBatteryAwareOperationRange2019a,kimMotionSpecificBatteryHealth2026}.
These developments provide important models for predicting energy consumption and battery evolution; however, they generally do not incorporate the complete closed-loop interaction among trajectory-dependent disturbance exposure, feedback-control and rotor demand, battery terminal-voltage evolution, and the resulting actuator authority directly within candidate-trajectory evaluation.
This distinction is important because the control actions required to execute
a trajectory determine the electrical load, while the evolving battery
terminal voltage can constrain the propulsion capability available to realize
that trajectory later in the mission.
Consequently, trajectories with comparable geometric length or flight time
can differ not only in energy consumption, but also in their closed-loop
tracking and actuator feasibility as the battery evolves.
The focus of the present work is therefore to incorporate this closed-loop trajectory--control--battery--actuator interaction directly into predictive trajectory selection under environmental disturbances.

Motivated by these considerations, this paper develops a battery-aware
predictive trajectory planning framework for multirotors operating under
environmental disturbances.
The central premise is that trajectory planning, closed-loop control, and
battery dynamics should not be treated independently: the selected trajectory
determines disturbance exposure and control demand, the resulting electrical
load governs battery evolution, and the battery terminal voltage in turn
constrains the available propulsion authority.
Accordingly, the main contributions of this paper are:
\vspace{-0.5\baselineskip}
\begin{enumerate} 
	\item A coupled closed-loop prediction framework that propagates vehicle and propulsion dynamics together with battery SOC, terminal voltage, electrical demand, and voltage-dependent rotor authority during candidate-trajectory evaluation.
	
	\item A battery-aware predictive trajectory planner that uses the coupled
	prediction to modify disturbance exposure while accounting for closed-loop
	tracking, mission energy, battery operating requirements, and
	battery-dependent actuator constraints.
	
	\item A closed-loop numerical evaluation over a $150$-s, $640$-m mission
	containing three localized disturbance regions, including comparison with the nominal trajectory, planner-ablation studies, a low-SOC actuator-stress case, and an execution-sensitivity study using four feedback-control architectures.
\end{enumerate}
%%%%%%%%%%%%%%%%%%%%%%%%%%%%%%%%%%%%%%%%%%%%%%%%%%%%%%%%%%%%%%%%%%%%%%%%%%%%%%%%
\section{System Modeling and Problem Formulation}
\label{sec:system_model}

%%%%%%%%%%%%%%%%%%%%%%%%%%%%%%%%%%%%%%%%%%%%%%%%%%%%%%%%%%%%%%%%%%%%%%%%%%%
\subsection{Multirotor Dynamics}
\label{subsec:quad_dynamics}

The multirotor is modeled as a rigid body with six degrees of freedom.
Let $\mathcal{F}_{I}=\{\mathbf{e}_{1},\mathbf{e}_{2},\mathbf{e}_{3}\}$
denote the inertial frame and
$\mathcal{F}_{B}=\{\mathbf{b}_{1},\mathbf{b}_{2},\mathbf{b}_{3}\}$
denote the body-fixed frame attached to the vehicle center of mass.
The inertial position and velocity of the vehicle are denoted by
$\mathbf{p}=[x,y,z]^{\top}\in\mathbb{R}^{3}$ and
$\mathbf{v}=[v_x,v_y,v_z]^{\top}\in\mathbb{R}^{3}$, respectively.
The body-fixed angular velocity is defined as
$\boldsymbol{\omega}=[p,q,r]^{\top}\in\mathbb{R}^{3}$, and
$\mathbf{R}\in\mathrm{SO}(3)$ denotes the rotation matrix from
$\mathcal{F}_{B}$ to $\mathcal{F}_{I}$.
The attitude is parameterized by the ZYX Euler-angle vector
$\boldsymbol{\Theta}=[\phi,\theta,\psi]^{\top}\in\mathbb{R}^{3}$, such that
\begin{equation}
	\mathbf{R}(\boldsymbol{\Theta})
	=
	\mathbf{R}_{z}(\psi)
	\mathbf{R}_{y}(\theta)
	\mathbf{R}_{x}(\phi).
	\label{eq:rotation_matrix}
\end{equation}
The corresponding rigid-body kinematics are
\begin{subequations}
	\label{eq:quad_kinematics}
	\begin{align}
		\dot{\mathbf{p}} &= \mathbf{v},
		\label{eq:position_kinematics}\\
		\dot{\mathbf{R}} &= \mathbf{R}[\boldsymbol{\omega}]_{\times},
		\label{eq:attitude_kinematics}
	\end{align}
\end{subequations}
where $[\cdot]_{\times}$ denotes the skew-symmetric operator. For
$\boldsymbol{\omega}=[p,q,r]^{\top}$,
\begin{equation}
	[\boldsymbol{\omega}]_{\times}
	=
	\begin{bmatrix}
		0  & -r & q \\
		r  & 0  & -p\\
		-q & p  & 0
	\end{bmatrix}.
	\label{eq:skew_operator}
\end{equation}
For implementation using Euler angles, the attitude kinematics can
equivalently be expressed as
\begin{equation}
	\dot{\boldsymbol{\Theta}}
	=
	\mathbf{T}(\boldsymbol{\Theta})\boldsymbol{\omega},
	\label{eq:euler_kinematics}
\end{equation}
where
\begin{equation}
	\mathbf{T}(\boldsymbol{\Theta})
	=
	\begin{bmatrix}
		1 & \sin\phi\tan\theta & \cos\phi\tan\theta\\
		0 & \cos\phi           & -\sin\phi\\
		0 & \sin\phi\sec\theta & \cos\phi\sec\theta
	\end{bmatrix},
	\label{eq:euler_rate_matrix}
\end{equation}
which is nonsingular for $|\theta|<\pi/2$.

Using the Newton--Euler formulation, the translational and rotational
dynamics are
\begin{subequations}
	\label{eq:quad_dynamics}
	\begin{align}
		m\dot{\mathbf{v}}
		&=
		-mg\mathbf{e}_{3}
		+
		\mathbf{R}\mathbf{e}_{3}f_T
		+
		\mathbf{f}_{\mathrm{air}},
		\label{eq:translational_dynamics}\\
		\mathbf{J}\dot{\boldsymbol{\omega}}
		&=
		-[\boldsymbol{\omega}]_{\times}
		\mathbf{J}\boldsymbol{\omega}
		+
		\boldsymbol{\tau}
		+
		\boldsymbol{\tau}_{g}
		+
		\boldsymbol{\tau}_{d}.
		\label{eq:rotational_dynamics}
	\end{align}
\end{subequations}
Here, $m\in\mathbb{R}_{>0}$ is the vehicle mass,
$g\in\mathbb{R}_{>0}$ is the gravitational acceleration, and
$\mathbf{e}_{3}=[0,0,1]^{\top}$ defines the upward inertial vertical axis.
The matrix
$\mathbf{J}=\mathrm{diag}(J_x,J_y,J_z)\in\mathbb{R}^{3\times3}$
is the positive-definite vehicle inertia matrix.
The scalar $f_T\in\mathbb{R}_{\geq0}$ denotes the total rotor thrust,
which acts along the body-fixed $\mathbf{b}_{3}$ axis, while
$\boldsymbol{\tau}
=
[\tau_{\phi},\tau_{\theta},\tau_{\psi}]^{\top}\in\mathbb{R}^{3}$
denotes the control torque generated by the rotors.
The term $\mathbf{f}_{\mathrm{air}}\in\mathbb{R}^{3}$ represents the
aerodynamic loading associated with the relative airflow between the
vehicle and the surrounding air.
The prescribed environmental disturbance enters this term through the
wind-velocity field introduced in Section~\ref{subsec:disturbance_model}.
The vector $\boldsymbol{\tau}_{d}\in\mathbb{R}^{3}$ denotes externally
induced disturbance moments. These moments are retained in the general
vehicle model, while the numerical study considers
$\boldsymbol{\tau}_{d}=\mathbf{0}$.

The gyroscopic torque induced by the rotating propellers is represented as
\begin{equation}
	\boldsymbol{\tau}_{g}
	=
	\boldsymbol{\omega}
	\times
	\left(
	J_r\mathbf{e}_{3}^{B}
	\sum_{i=1}^{4}\sigma_i\omega_i
	\right),
	\label{eq:gyro_torque}
\end{equation}
where $J_r\in\mathbb{R}_{>0}$ denotes the rotor polar moment of inertia,
$\mathbf{e}_{3}^{B}=[0,0,1]^{\top}$ denotes the rotor-axis unit vector
expressed in $\mathcal{F}_{B}$, and
$\sigma_i\in\{-1,+1\}$ specifies the rotation direction of the $i$th rotor.
The distinction between $\mathbf{e}_{3}$ and $\mathbf{e}_{3}^{B}$ is made
explicit because the former denotes the inertial vertical direction,
whereas the latter denotes the rotor axis expressed in body coordinates.

The total thrust is related to the individual rotor speeds according to
\begin{equation}
	f_T
	=
	k_T\sum_{i=1}^{4}\omega_i^2,
	\label{eq:total_thrust}
\end{equation}
where $k_T\in\mathbb{R}_{>0}$ is the rotor thrust coefficient and
$\omega_i\in\mathbb{R}_{\geq0}$ is the angular speed of the $i$th rotor.
For compactness, the rigid-body vehicle state is defined as
\begin{equation}
	\mathbf{x}
	=
	\begin{bmatrix}
		\mathbf{p}^{\top} &
		\mathbf{v}^{\top} &
		\boldsymbol{\Theta}^{\top} &
		\boldsymbol{\omega}^{\top}
	\end{bmatrix}^{\top}
	\in\mathbb{R}^{12},
	\label{eq:quad_state}
\end{equation}
with generalized control input
\begin{equation}
	\mathbf{u}
	=
	\begin{bmatrix}
		f_T &
		\boldsymbol{\tau}^{\top}
	\end{bmatrix}^{\top}
	\in\mathbb{R}^{4}.
	\label{eq:quad_input}
\end{equation}
The model in \eqref{eq:quad_dynamics} explicitly retains the coupling
between translational and rotational motion through
$\mathbf{R}\mathbf{e}_{3}f_T$. Consequently, lateral acceleration cannot
be generated independently of vehicle attitude, reflecting the
underactuated nature of the multirotor.
This coupling is important in the present work because a candidate
trajectory determines the required vehicle attitude and thrust history,
which subsequently determines the rotor demand and associated battery
load.

%%%%%%%%%%%%%%%%%%%%%%%%%%%%%%%%%%%%%%%%%%%%%%%%%%%%%%%%%%%%%%%%%%%%%%%%%%%
\subsection{Motor and Propulsion Dynamics}
\label{subsec:propulsion_dynamics}

The generalized control input in \eqref{eq:quad_input} is generated by four
rotors with angular velocities
$\boldsymbol{\omega}_{r}
=
[\omega_1,\omega_2,\omega_3,\omega_4]^{\top}
\in\mathbb{R}_{\geq0}^{4}$.
For the ``$+$'' configuration considered here, rotors $1$--$4$ are located
along the $+\mathbf{b}_1$, $+\mathbf{b}_2$, $-\mathbf{b}_1$, and
$-\mathbf{b}_2$ axes, respectively.
Rotors $1$ and $3$ rotate in one direction and rotors $2$ and $4$ rotate
in the opposite direction, such that
$\sigma_1=\sigma_3=+1$ and $\sigma_2=\sigma_4=-1$ under the adopted yaw
moment convention.
The thrust and control moments are related to the squared rotor speeds
according to
\begin{equation}
	\begin{bmatrix}
		f_T\\
		\tau_{\phi}\\
		\tau_{\theta}\\
		\tau_{\psi}
	\end{bmatrix}
	=
%	\underbrace{
		\begin{bmatrix}
			k_T & k_T & k_T & k_T\\
			0 & lk_T & 0 & -lk_T\\
			-lk_T & 0 & lk_T & 0\\
			k_Q & -k_Q & k_Q & -k_Q
		\end{bmatrix}
%	}_{\mathbf{B}_{\omega}}
	\begin{bmatrix}
		\omega_1^2\\
		\omega_2^2\\
		\omega_3^2\\
		\omega_4^2
	\end{bmatrix},
	\label{eq:rotor_allocation}
\end{equation}
where $l\in\mathbb{R}_{>0}$ is the distance from each rotor to the vehicle
center of mass, and $k_T,k_Q\in\mathbb{R}_{>0}$ are the thrust and reaction
torque coefficients, respectively.
The controller-requested squared rotor speeds are obtained from the inverse
allocation as
\begin{equation}
	\boldsymbol{\nu}_{\mathrm{req}}
	=
	\max
	\left\{
	\mathbf{0},
	\mathbf{B}_{\omega}^{-1}\mathbf{u}
	\right\},
	\qquad
	\boldsymbol{\omega}_{r,\mathrm{req}}
	=
	\sqrt{\boldsymbol{\nu}_{\mathrm{req}}},
	\label{eq:requested_rotor_speed}
\end{equation}
where the maximum and square-root operations are applied element-wise.
Thus, negative squared-speed values resulting from the inverse allocation
are clipped to zero before the corresponding rotor-speed request is formed.
The requested speeds are subsequently bounded by the nominal motor-speed
capability,
\begin{equation}
	\bar{\omega}_{i,\mathrm{req}}
	=
	\min
	\left\{
	\omega_{i,\mathrm{req}},
	\omega_{\max}^{\mathrm{nom}}
	\right\},
	\qquad
	i\in\{1,\ldots,4\},
	\label{eq:nominal_rotor_limit}
\end{equation}
before application of the battery-dependent actuator limit.
The resulting commanded rotor speed is
\begin{equation}
	\omega_{i,\mathrm{cmd}}
	=
	\min
	\left\{
	\bar{\omega}_{i,\mathrm{req}},
	\omega_{\max}(V_b)
	\right\},
	\qquad
	i\in\{1,\ldots,4\},
	\label{eq:commanded_rotor_speed}
\end{equation}
where $\omega_{\max}(V_b)$ denotes the instantaneous rotor-speed capability
associated with the battery terminal voltage and is developed in
Section~III.
By construction,
\begin{equation}
	\omega_{\max}(V_b)
	\leq
	\omega_{\max}^{\mathrm{nom}},
	\label{eq:battery_limit_bounded}
\end{equation}
such that battery discharge can only reduce the available rotor-speed
capability relative to the nominal hardware limit.
The actual rotor dynamics are represented using the first-order model
\begin{equation}
	\dot{\omega}_i
	=
	\frac{1}{\tau_m}
	\left(
	\omega_{i,\mathrm{cmd}}-\omega_i
	\right),
	\qquad
	i\in\{1,\ldots,4\},
	\label{eq:motor_dynamics}
\end{equation}
where $\tau_m\in\mathbb{R}_{>0}$ denotes the motor time constant.
Equations~\eqref{eq:requested_rotor_speed}--\eqref{eq:motor_dynamics}
therefore distinguish the controller-requested rotor speed, the
battery-limited motor command, and the resulting physical rotor speed.
This distinction is central to assessing whether the requested control
action remains realizable as the battery terminal voltage evolves.

%%%%%%%%%%%%%%%%%%%%%%%%%%%%%%%%%%%%%%%%%%%%%%%%%%%%%%%%%%%%%%%%%%%%%%%%%%%
\subsection{Disturbance Model}
\label{subsec:disturbance_model}

The vehicle is assumed to operate in an environment containing spatially
localized wind disturbances.
For the present formulation, the disturbance field is assumed to be known
a priori over the mission horizon.
Thus, the planner is supplied with the locations and prescribed wind
characteristics of the disturbance regions; uncertainty in this
disturbance information is not considered in the present study.
Let
$\mathbf{p}_{h}=[x,y]^{\top}\in\mathbb{R}^{2}$
denote the horizontal vehicle position.
The external environmental disturbance is represented by the wind-velocity
field
\begin{equation}
	\mathbf{v}_{w}(\mathbf{p})
	=
	\sum_{j=1}^{N_d}
	\mathbf{v}_{w,j}
	\exp
	\left[
	-\frac{(z-z_{d,j})^2}
	{2\sigma_{d,j}^{2}}
	\right]
	\chi_j(\mathbf{p}_{h}),
	\label{eq:wind_field}
\end{equation}
where $N_d$ is the number of disturbance regions,
$\mathbf{v}_{w,j}\in\mathbb{R}^{3}$ denotes the prescribed wind velocity
associated with the $j$th region, and $z_{d,j}$ and
$\sigma_{d,j}\in\mathbb{R}_{>0}$ define its vertical center and spatial
spread, respectively.
The indicator
$\chi_j(\mathbf{p}_{h})\in\{0,1\}$
identifies whether the vehicle lies within the horizontal extent of the
$j$th disturbance region.
For the mission considered in this work, the horizontal reference path is
prescribed and only the vertical trajectory is modified by the planner.
Consequently, the horizontal disturbance-region encounter intervals remain
fixed during candidate-trajectory evaluation despite the binary region
indicator.

The aerodynamic effect of the external disturbance is incorporated through
the relative air velocity
\begin{equation}
	\mathbf{v}_{\mathrm{rel}}
	=
	\mathbf{v}
	-
	\mathbf{v}_{w}(\mathbf{p}),
	\label{eq:relative_air_velocity}
\end{equation}
and the corresponding aerodynamic force in
\eqref{eq:translational_dynamics} is modeled as
\begin{equation}
	\mathbf{f}_{\mathrm{air}}
	=
	-\mathbf{D}
	\left(
	\mathbf{v}_{\mathrm{rel}}
	\odot
	|\mathbf{v}_{\mathrm{rel}}|
	\right),
	\label{eq:wind_force}
\end{equation}
where $\odot$ denotes element-wise multiplication and
\begin{equation}
	\mathbf{D}
	=
	\operatorname{diag}
	\left(
	d_x,d_y,d_z
	\right)
	\succ\mathbf{0}
	\label{eq:drag_matrix}
\end{equation}
contains the effective direction-dependent quadratic drag coefficients.
Accordingly, $\mathbf{v}_{w}(\mathbf{p})$ represents the prescribed
external environmental disturbance, whereas
$\mathbf{f}_{\mathrm{air}}$ represents the resulting aerodynamic loading
on the vehicle.
When $\mathbf{v}_{w}=\mathbf{0}$,
\eqref{eq:wind_force} reduces to the aerodynamic resistance associated with
vehicle motion through still air; within a disturbance region, the
prescribed wind modifies the relative airflow and therefore the resulting
aerodynamic load.
The selected trajectory consequently determines the altitude at which each
disturbance region is encountered and, through
\eqref{eq:wind_field}--\eqref{eq:wind_force}, the aerodynamic loading that
must be rejected by the closed-loop controller.

%%%%%%%%%%%%%%%%%%%%%%%%%%%%%%%%%%%%%%%%%%%%%%%%%%%%%%%%%%%%%%%%%%%%%%%%%%%
\subsection{Problem Formulation}
\label{subsec:problem_formulation}

Consider the multirotor dynamics in
\eqref{eq:quad_dynamics}--\eqref{eq:motor_dynamics}
executing a nominal mission trajectory
$\mathbf{p}_{0}(t)$ through the prescribed disturbance field in
\eqref{eq:wind_field}.
The objective is to determine a modified reference trajectory
$\mathbf{p}_{d}(t;\boldsymbol{\theta})$, parameterized by the planning
variables $\boldsymbol{\theta}$, while accounting for the prescribed
mission requirements and the closed-loop energetic and actuator
consequences of the selected trajectory.
For each candidate trajectory, the coupled closed-loop vehicle, motor, and
battery dynamics are propagated over the mission horizon according to the
following causal and algebraic dependencies:
\begin{equation}
	\begin{aligned}
		\mathbf{p}_{d}
		&\rightarrow
		\mathbf{v}_{w}
		\rightarrow
		\mathbf{f}_{\mathrm{air}}
		\rightarrow
		\{\mathbf{x},\mathbf{u}\}
		\rightarrow
		\boldsymbol{\omega}_{r,\mathrm{req}}
		\rightarrow
		\boldsymbol{\omega}_{r,\mathrm{cmd}}
		\\
		&\rightarrow
		\boldsymbol{\omega}_{r}
		\rightarrow
		P_b
		\rightarrow
		\left\{
		\mathrm{SOC},
		V_b,
		\omega_{\max}
		\right\},
	\end{aligned}
	\label{eq:coupled_prediction_chain}
\end{equation}
where $P_b$ denotes the total electrical power drawn from the battery.
The terminal voltage subsequently determines
$\omega_{\max}(V_b)$ and therefore feeds back into the commanded rotor
speed in \eqref{eq:commanded_rotor_speed}, producing the algebraic
battery--actuator coupling developed in Section~III.

The resulting battery state and actuator authority are used to assess the
mission-level operating requirements
\begin{subequations}
	\label{eq:feasibility_constraints}
	\begin{align}
		\mathrm{SOC}(t)
		&\geq
		\mathrm{SOC}_{\min},
		\label{eq:soc_feasibility}\\
		V_b(t)
		&\geq
		V_{\min},
		\label{eq:voltage_feasibility}\\
		\omega_{i,\mathrm{req}}(t)
		&\leq
		\omega_{\max}^{\mathrm{nom}},
		\qquad
		i\in\{1,\ldots,4\},
		\label{eq:nominal_actuator_feasibility}\\
		\bar{\omega}_{i,\mathrm{req}}(t)
		&\leq
		\omega_{\max}\!\left(V_b(t)\right),
		\qquad
		i\in\{1,\ldots,4\}.
		\label{eq:battery_actuator_feasibility}
	\end{align}
\end{subequations}
Equation~\eqref{eq:nominal_actuator_feasibility} verifies that the original
controller request remains within the nominal motor-speed capability,
whereas \eqref{eq:battery_actuator_feasibility} evaluates whether the
nominally admissible request remains realizable under the instantaneous
battery-dependent actuator limit.
Because
$\omega_{\max}(V_b)\leq\omega_{\max}^{\mathrm{nom}}$ by
\eqref{eq:battery_limit_bounded}, satisfaction of both conditions ensures
that neither nominal nor battery-dependent rotor-speed saturation is
required during trajectory execution.
Importantly, actuator feasibility is assessed using rotor-speed requests
prior to the corresponding saturation operation; otherwise, feasibility
would be satisfied by construction.

In the trajectory optimization developed in Section~IV, the battery and
actuator operating requirements in
\eqref{eq:feasibility_constraints} are incorporated through penalty terms,
whereas the planning-variable bounds constitute the hard optimization
constraints.
Mission feasibility is subsequently verified using
\eqref{eq:feasibility_constraints} after optimization.
Here, ``predictive'' refers to evaluating candidate trajectories through
forward closed-loop propagation of the coupled
vehicle--motor--battery model over the complete mission horizon.
The present implementation performs this mission-level optimization prior
to execution rather than in a receding-horizon manner.
The battery dynamics, battery-dependent actuator authority, and
corresponding trajectory optimization are developed in  Sections~\ref{sec:battery_actuator}
and~\ref{sec:predictive_planner}, respectively.
%%%%%%%%%%%%%%%%%%%%%%%%%%%%%%%%%%%%%%%%%%%%%%%%%%%%%%%%%%%%%%%%%%%%%%%%%%%%%%%%
\section{Battery and Actuator Model}
\label{sec:battery_actuator}

The energetic and actuator feasibility of a candidate trajectory depends not
only on the total energy available onboard, but also on the instantaneous
electrical load, battery state, and corresponding propulsion capability.
To capture these interactions during candidate-trajectory prediction, an
SOC-dependent equivalent-circuit battery model is coupled with the propulsion
power demand and a terminal-voltage-dependent rotor-speed limit.
This formulation allows battery charge, transient voltage response, electrical
power capability, and available actuator authority to evolve simultaneously
with the closed-loop vehicle and motor dynamics.

%%%%%%%%%%%%%%%%%%%%%%%%%%%%%%%%%%%%%%%%%%%%%%%%%%%%%%%%%%%%%%%%%%%%%%%%%%%
\subsection{SOC-Dependent Equivalent-Circuit Battery Model}
\label{subsec:battery_model}

The battery pack is represented by a two-RC Thevenin equivalent circuit with
SOC-dependent parameters.
The model consists of an SOC-dependent open-circuit voltage (OCV), an
SOC-dependent ohmic resistance $R_0(s)$, and two polarization branches
characterized by $R_j(s)$ and $C_j(s)$, $j\in\{1,2\}$.
The two polarization states provide distinct transient relaxation time scales
while retaining a sufficiently low-order representation for repeated
candidate-trajectory propagation.

Let $s\in[0,1]$ denote the normalized battery state of charge (SOC),
$I_b\geq0$ the pack discharge current, and $V_{p,1}$ and $V_{p,2}$ the
polarization voltages.
The battery state is
\begin{equation}
	\mathbf{x}_b
	=
	\begin{bmatrix}
		s & V_{p,1} & V_{p,2}
	\end{bmatrix}^{\top},
	\label{eq:battery_state_vector}
\end{equation}
with dynamics
\begin{subequations}
	\label{eq:battery_states}
	\begin{align}
		\dot{s}
		&=
		-\frac{I_b}{3600Q_b},
		\label{eq:soc_dynamics}\\
		\dot{V}_{p,1}
		&=
		\frac{I_b}{C_1(s)}
		-
		\frac{V_{p,1}}{R_1(s)C_1(s)},
		\label{eq:polarization_dynamics_1}\\
		\dot{V}_{p,2}
		&=
		\frac{I_b}{C_2(s)}
		-
		\frac{V_{p,2}}{R_2(s)C_2(s)},
		\label{eq:polarization_dynamics_2}
	\end{align}
\end{subequations}
where $Q_b$ is the rated battery capacity in Ah.
The factor $3600$ converts ampere-hours to ampere-seconds, and the sign
convention $I_b>0$ corresponds to battery discharge.
The resistive and polarization parameters
$R_0(s)$, $R_1(s)$, $R_2(s)$, $C_1(s)$, and $C_2(s)$ are evaluated from
SOC-dependent lookup tables using piecewise-linear interpolation over the
prescribed operating SOC range.

The open-circuit voltage is represented as a nonlinear function of SOC.
Analytical and tabulated OCV--SOC representations are commonly obtained from
battery characterization data
\cite{chenOnlineStateCharge2019,
	wengUnifiedOpencircuitvoltageModel2014,
	pattipatiOpenCircuitVoltage2014}.
For the present model, the cell-level OCV relationship is
\begin{equation}
	V_{\mathrm{oc},c}(s)
	=
	a_0+a_1s-a_2e^{-a_3s},
	\label{eq:cell_ocv}
\end{equation}
and the corresponding pack-level OCV for $N_s$ series-connected cells is
\begin{equation}
	V_{\mathrm{oc}}(s)
	=
	N_sV_{\mathrm{oc},c}(s).
	\label{eq:pack_ocv}
\end{equation}
The resistive and polarization parameters
$R_0(s)$, $R_1(s)$, $R_2(s)$, $C_1(s)$, and $C_2(s)$ are evaluated as
SOC-dependent lookup quantities over the operating SOC range.
Equivalently, the polarization capacitances may be defined from the
SOC-dependent time constants according to
\begin{equation}
	C_j(s)
	=
	\frac{\tau_j(s)}{R_j(s)},
	\qquad j\in\{1,2\}.
	\label{eq:battery_rc_time_constants}
\end{equation}

The battery terminal voltage is
\begin{equation}
	V_b
	=
	V_{\mathrm{oc}}(s)
	-
	V_{p,1}
	-
	V_{p,2}
	-
	R_0(s)I_b.
	\label{eq:battery_terminal_voltage}
\end{equation}
Since the propulsion model specifies the required electrical load through
battery power,
\begin{equation}
	P_b=V_bI_b,
	\label{eq:battery_power_identity}
\end{equation}
the battery current must be determined consistently with the instantaneous
terminal voltage.
Defining the voltage prior to the instantaneous ohmic drop as
\begin{equation}
	\bar{V}_b
	=
	V_{\mathrm{oc}}(s)
	-
	V_{p,1}
	-
	V_{p,2},
	\label{eq:available_battery_voltage}
\end{equation}
substitution of \eqref{eq:battery_terminal_voltage} into
\eqref{eq:battery_power_identity} gives
\begin{equation}
	R_0(s)I_b^2
	-
	\bar{V}_b I_b
	+
	P_b
	=
	0.
	\label{eq:battery_current_quadratic}
\end{equation}
The physically relevant low-current discharge solution is therefore
\begin{equation}
	I_b
	=
	\frac{
		\bar{V}_b
		-
		\sqrt{
			\bar{V}_b^{\,2}
			-
			4R_0(s)P_b
		}
	}{
		2R_0(s)
	}.
	\label{eq:battery_current}
\end{equation}
A real-valued solution requires
\begin{equation}
	\bar{V}_b^{\,2}
	-
	4R_0(s)P_b
	\geq0,
	\label{eq:battery_discriminant_constraint}
\end{equation}
or, equivalently,
\begin{equation}
	P_b
	\leq
	P_{b,\max}
	\triangleq
	\frac{\bar{V}_b^{\,2}}{4R_0(s)}.
	\label{eq:battery_power_limit}
\end{equation}
Equation~\eqref{eq:battery_power_limit} defines the instantaneous
power-delivery condition associated with the equivalent-circuit model.
A candidate trajectory violating this condition is classified as
electrically infeasible during candidate-trajectory evaluation.
Consequently, the battery model captures both cumulative charge depletion and
the transient voltage depression produced by the closed-loop electrical load,
while allowing the instantaneous power capability to vary with battery state.

The model assumes discharge-only operation and neglects thermal dynamics,
ageing, cell-to-cell imbalance, hysteresis, self-discharge, and regenerative
charging.
The SOC-dependent equivalent-circuit parameters used in the present numerical
study are prescribed lookup quantities constructed from the baseline
single-RC parameterization to introduce SOC dependence and a second
polarization time scale.
They are therefore used as a mission-level model-form parameterization and
should not be interpreted as parameters identified from cell-specific
experimental characterization.

%%%%%%%%%%%%%%%%%%%%%%%%%%%%%%%%%%%%%%%%%%%%%%%%%%%%%%%%%%%%%%%%%%%%%%%%%%%
\subsection{Propulsion Power and Battery Load}
\label{subsec:battery_load}

The electrical load imposed on the battery is determined from the individual
rotor states rather than from a prescribed mission-level energy rate.
Rotorcraft power consumption depends strongly on propulsion demand and
operating condition
\cite{gongModelingPowerConsumptions2023,
	zhangEnergyConsumptionModels2021}.
Moreover, rotor acceleration introduces an additional transient load because
the motor must accelerate the rotating motor--propeller assembly
\cite{jacewiczQuadrotorModelEnergy2022}.
To retain these effects without introducing a detailed motor electrical model
into each candidate-trajectory rollout, the electrical power associated with
the $i$th motor is approximated as
\begin{equation}
	P_{m,i}
	=
	\frac{k_P}{\eta_m}\omega_i^3
	+
	k_a
	\left[
	\max
	\left(
	\omega_{i,\mathrm{cmd}}-\omega_i,0
	\right)
	\right]^2
	+
	P_{\mathrm{idle}},
	\label{eq:motor_power}
\end{equation}
where $k_P$ is the propeller power coefficient,
$\eta_m\in(0,1]$ is the effective motor--propulsion efficiency,
$k_a$ is an empirical transient-load coefficient, and
$P_{\mathrm{idle}}$ denotes the per-motor idle electrical power.
The first term captures the cubic dependence of rotor aerodynamic power on
angular speed, whereas the second term approximates the additional electrical
demand associated with increasing rotor speed.
Because regenerative operation is not modeled, the transient correction is
applied only when $\omega_{i,\mathrm{cmd}}>\omega_i$.
Using the first-order motor dynamics in \eqref{eq:motor_dynamics},
\begin{equation}
	\omega_{i,\mathrm{cmd}}-\omega_i
	=
	\tau_m\dot{\omega}_i,
	\label{eq:motor_error_acceleration_relation}
\end{equation}
so that the transient correction increases with positive rotor acceleration.
Equation~\eqref{eq:motor_power} is therefore used as a reduced-order
propulsion-load approximation rather than as a detailed electromechanical
BLDC motor model.

The total electrical power requested from the battery is
\begin{equation}
	P_b
	=
	\sum_{i=1}^{4}P_{m,i}
	+
	P_{\mathrm{aux}},
	\label{eq:battery_power}
\end{equation}
where $P_{\mathrm{aux}}$ represents the auxiliary electrical load associated
with onboard electronics and other non-propulsive systems.
The inclusion of a non-propulsive load is consistent with energy-consumption
models that distinguish propulsion and onboard-system power demand
\cite{jacewiczQuadrotorModelEnergy2022,
	zhangEnergyConsumptionModels2021}.
The cumulative electrical energy consumed over $[0,t]$ is
\begin{equation}
	E_b(t)
	=
	\frac{1}{3600}
	\int_{0}^{t}P_b(\tau)\,d\tau
	\qquad [\mathrm{Wh}].
	\label{eq:battery_energy}
\end{equation}
Thus, trajectory tracking and environmental disturbances modify the
feedback-control demand and rotor-speed histories, which subsequently alter
the electrical load, battery state, and available propulsion capability.

%%%%%%%%%%%%%%%%%%%%%%%%%%%%%%%%%%%%%%%%%%%%%%%%%%%%%%%%%%%%%%%%%%%%%%%%%%%
\subsection{Battery-Dependent Actuator Authority}
\label{subsec:battery_actuator_authority}

The battery and propulsion models are coupled by allowing the maximum
achievable rotor speed to vary with the instantaneous battery terminal
voltage.
For a voltage-limited electric propulsion system, the achievable motor speed
is coupled to the available supply voltage through the motor back-EMF
relationship.
Rather than introducing a detailed motor torque--speed and electronic-speed-
controller model into each candidate-trajectory rollout, this dependence is
represented using a reduced-order voltage-dependent actuator envelope.

Let $\omega_{\max}^{\mathrm{nom}}$ denote the nominal rotor-speed capability
and $V_{\mathrm{ref}}$ the pack terminal voltage at which this nominal
rotor-speed capability is available.
The available rotor-speed limit is modeled as
\begin{equation}
	\omega_{\max}(V_b)
	=
	\omega_{\max}^{\mathrm{nom}}
	\min
	\left\{
	1,
	\frac{V_b}{V_{\mathrm{ref}}}
	\right\},
	\qquad
	V_b\geq V_{\min}.
	\label{eq:voltage_dependent_speed}
\end{equation}
By construction,
\begin{equation}
	\omega_{\max}(V_b)
	\leq
	\omega_{\max}^{\mathrm{nom}},
	\label{eq:voltage_speed_bound}
\end{equation}
consistent with the actuator hierarchy introduced in
Section~\ref{subsec:propulsion_dynamics}.
The numerical values of $V_{\mathrm{ref}}$ and
$\omega_{\max}^{\mathrm{nom}}$ are specified with the propulsion-system
parameters in the numerical study.

Substitution of \eqref{eq:voltage_dependent_speed} into the rotor-command
constraint gives
\begin{equation}
	\omega_{i,\mathrm{cmd}}
	=
	\min
	\left\{
	\bar{\omega}_{i,\mathrm{req}},
	\omega_{\max}(V_b)
	\right\},
	\qquad
	i\in\{1,\ldots,4\},
	\label{eq:battery_limited_motor_command}
\end{equation}
where $\bar{\omega}_{i,\mathrm{req}}$ is the nominally admissible rotor-speed
request defined in \eqref{eq:nominal_rotor_limit}.
Accordingly, the raw controller request $\omega_{i,\mathrm{req}}$, nominally
admissible request $\bar{\omega}_{i,\mathrm{req}}$, battery-limited command
$\omega_{i,\mathrm{cmd}}$, and actual rotor speed $\omega_i$ remain distinct
throughout the coupled prediction model.

To quantify the remaining battery-dependent propulsion reserve, define
\begin{subequations}
	\label{eq:authority_metrics}
	\begin{align}
		\Delta\omega
		&=
		\omega_{\max}(V_b)
		-
		\max_i\bar{\omega}_{i,\mathrm{req}},
		\label{eq:authority_margin}\\
		\eta_{\omega}
		&=
		\frac{
			\max_i\bar{\omega}_{i,\mathrm{req}}
		}{
			\omega_{\max}(V_b)
		}.
		\label{eq:authority_utilization}
	\end{align}
\end{subequations}
Here, $\Delta\omega$ is the instantaneous rotor-speed reserve and
$\eta_{\omega}$ is the fraction of the battery-dependent rotor-speed
capability requested by the controller.
Thus, $\Delta\omega>0$ and $\eta_{\omega}<1$ indicate available actuator
reserve, whereas $\eta_{\omega}\geq1$ indicates that the nominally admissible
request has reached or exceeded the instantaneous battery-dependent
capability.
Overall actuator feasibility additionally requires the unsaturated rotor
request to satisfy the nominal hardware condition in
\eqref{eq:nominal_actuator_feasibility}.

Importantly, the terminal voltage and available actuator capability cannot be
evaluated independently.
The terminal voltage determines $\omega_{\max}$, which constrains the rotor
commands and therefore changes the electrical power demand; the resulting
battery current, in turn, determines the terminal voltage.
Accordingly, at each prediction step the coupled algebraic dependence
\begin{equation}
	\begin{aligned}
		V_b
		&\rightarrow
		\omega_{\max}
		\rightarrow
		\boldsymbol{\omega}_{r,\mathrm{cmd}}
		\rightarrow
		P_b
		\rightarrow
		I_b
		\rightarrow
		V_b
	\end{aligned}
	\label{eq:battery_actuator_coupling}
\end{equation}
is resolved iteratively to a prescribed terminal-voltage tolerance.
During this iteration, the differential states
$s$, $V_{p,1}$, $V_{p,2}$, and $\boldsymbol{\omega}_r$ are held fixed,
whereas $V_b$, $\omega_{\max}$,
$\boldsymbol{\omega}_{r,\mathrm{cmd}}$, $P_b$, and $I_b$ are updated.
After convergence, the resulting current is used in
\eqref{eq:battery_states}, and the converged rotor command is used in
\eqref{eq:motor_dynamics}, to propagate the battery and motor states.

This coupled representation enables each candidate trajectory to be evaluated
through the sequence
\begin{equation}
	\begin{aligned}
		p_d
		&\rightarrow v_w
		\rightarrow \{x,u\}
		\rightarrow \boldsymbol{\omega}_{r,\mathrm{req}}
		\rightarrow \boldsymbol{\omega}_{r,\mathrm{cmd}}
		\rightarrow \boldsymbol{\omega}_r \\
		&\rightarrow P_b
		\rightarrow
		\{I_b,V_b,s,V_{p,1},V_{p,2}\}
		\rightarrow \omega_{\max},
	\end{aligned}
	\label{eq:coupled_prediction_chain}
\end{equation}
thereby allowing the planner to account simultaneously for disturbance-induced
control demand, electrical energy consumption, battery evolution,
instantaneous power feasibility, and battery-dependent actuator authority.
%%%%%%%%%%%%%%%%%%%%%%%%%%%%%%%%%%%%%%%%%%%%%%%%%%%%%%%%%%%%%%%%%%%%%%%%%%%%%%%%
\section{Battery-Aware Predictive Trajectory Planning}
\label{sec:predictive_planner}

The proposed planner modifies a nominal mission trajectory prior to execution
such that the anticipated energetic, tracking, battery, and actuator
consequences of environmental disturbances are explicitly considered.
Rather than optimizing an unconstrained three-dimensional trajectory, the
planner employs a low-dimensional parameterization that preserves the nominal
horizontal mission while allowing localized altitude modifications in the
vicinity of known disturbance regions.
Each candidate trajectory is evaluated through closed-loop propagation of the
coupled vehicle, motor, battery, and actuator dynamics developed in
Sections~II and~III.
Consequently, the planning decision depends not only on the geometric
trajectory modification, but also on the feedback-control demand, electrical
load, battery evolution, and available actuator authority required to execute
the candidate trajectory.

%%%%%%%%%%%%%%%%%%%%%%%%%%%%%%%%%%%%%%%%%%%%%%%%%%%%%%%%%%%%%%%%%%%%%%%%%%%
\subsection{Predictive Planning Framework}
\label{subsec:predictive_framework}

Let $\mathbf{p}_0(t)$ denote the nominal mission trajectory and let
$\boldsymbol{\delta}
=[\delta_1,\ldots,\delta_{N_d}]^{\top}$ denote the planning vector, where
$\delta_j$ specifies the signed altitude modification associated with the
$j$th disturbance region.
The battery-aware trajectory is obtained from
\begin{equation}
	\boldsymbol{\delta}^{*}
	=
	\arg\min_{\boldsymbol{\delta}\in\mathcal{D}}
	J(\boldsymbol{\delta}),
	\label{eq:planner_optimization}
\end{equation}
where the box-constrained planning domain is
\begin{equation}
	\mathcal{D}_{\mathrm{box}}
	=
	\left\{
	\boldsymbol{\delta}\in\mathbb{R}^{N_d}:
	-\delta_{\max}\leq\delta_j\leq\delta_{\max},
	\;j=1,\ldots,N_d
	\right\}.
	\label{eq:planner_box_bounds}
\end{equation}
The signed variables allow the optimizer to select either upward or downward
altitude modifications according to the predicted consequences of each
candidate trajectory.
The complete admissible set is
\begin{equation}
	\mathcal{D}
	=
	\left\{
	\boldsymbol{\delta}\in\mathcal{D}_{\mathrm{box}}:
	\begin{array}{l}
		z_{\min}\leq z_d(t;\boldsymbol{\delta})\leq z_{\max},\\
		|\dot z_d(t;\boldsymbol{\delta})|\leq v_{z,\max},\\
		|\ddot z_d(t;\boldsymbol{\delta})|\leq a_{z,\max},
		\quad t\in[0,T_f]
	\end{array}
	\right\}.
	\label{eq:planner_admissible_set}
\end{equation}
Thus, altitude, vertical-velocity, and vertical-acceleration requirements are
enforced directly during trajectory optimization.

For each $\boldsymbol{\delta}$, the complete mission is propagated using a
prediction time step $\Delta t_p$.
The reference determines disturbance exposure and closed-loop control demand,
which subsequently determine the requested and achievable rotor speeds,
electrical load, battery evolution, and available actuator authority.
The principal dependencies during candidate evaluation are summarized as
\begin{equation}
	\begin{aligned}
		\mathbf{p}_d
		&\rightarrow
		\mathbf{v}_w
		\rightarrow
		\{\mathbf{x},\mathbf{u}\}
		\rightarrow
		\boldsymbol{\omega}_{r,\mathrm{req}}
		\rightarrow
		\bar{\boldsymbol{\omega}}_{r,\mathrm{req}}
		\rightarrow
		\boldsymbol{\omega}_{r,\mathrm{cmd}}
		\\
		&\rightarrow
		\boldsymbol{\omega}_r
		\rightarrow
		P_b
		\rightarrow
		\{I_b,V_b,s,V_{p,1},V_{p,2}\},
	\end{aligned}
	\label{eq:planner_prediction_chain}
\end{equation}
where $V_b$ additionally determines $\omega_{\max}(V_b)$ and therefore feeds
back algebraically into $\boldsymbol{\omega}_{r,\mathrm{cmd}}$.
Here, predictive refers to forward propagation of candidate trajectories
through the coupled closed-loop model over the complete mission horizon.
The optimization is performed prior to mission execution rather than in a
receding-horizon manner.

%%%%%%%%%%%%%%%%%%%%%%%%%%%%%%%%%%%%%%%%%%%%%%%%%%%%%%%%%%%%%%%%%%%%%%%%%%%
\subsection{Candidate Trajectory Parameterization}
\label{subsec:trajectory_parameterization}

To preserve the prescribed horizontal mission while modifying disturbance
exposure, the candidate reference is constructed by superimposing localized
vertical deviations onto the nominal trajectory:
\begin{equation}
	\mathbf{p}_d(t;\boldsymbol{\delta})
	=
	\mathbf{p}_0(t)
	+
	\mathbf{e}_3
	\sum_{j=1}^{N_d}
	\delta_j b_j(t),
	\label{eq:candidate_trajectory}
\end{equation}
where $b_j(t)$ is a compact $C^2$ basis function associated with disturbance
window $j$.
Let $[t_j^-,t_j^+]$ denote the corresponding nominal disturbance-encounter
interval and let $T_{\mathrm{tr}}>0$ denote the transition time.
Defining
\begin{equation}
	q_j(t)
	=
	\operatorname{sat}_{[0,1]}
	\left(
	\frac{t-(t_j^- -T_{\mathrm{tr}})}
	{t_j^+-t_j^-+2T_{\mathrm{tr}}}
	\right),
	\label{eq:bump_coordinate}
\end{equation}
the basis function used in the planner is
\begin{equation}
	b_j(t)
	=
	64q_j^3(t)\left[1-q_j(t)\right]^3.
	\label{eq:trajectory_bump}
\end{equation}
This construction satisfies $0\leq b_j(t)\leq1$, with zero value, velocity,
and acceleration at the boundaries of its support.
The basis reaches $b_j=1$ at the center of the corresponding interval, such
that $\delta_j$ represents the signed peak altitude modification associated
with the $j$th disturbance region.

The corresponding reference velocity and acceleration are
\begin{subequations}
	\label{eq:candidate_derivatives}
	\begin{align}
		\dot{\mathbf{p}}_d
		&=
		\dot{\mathbf{p}}_0+
		\mathbf{e}_3
		\sum_{j=1}^{N_d}\delta_j\dot b_j,
		\\
		\ddot{\mathbf{p}}_d
		&=
		\ddot{\mathbf{p}}_0+
		\mathbf{e}_3
		\sum_{j=1}^{N_d}\delta_j\ddot b_j .
	\end{align}
\end{subequations}
The transition duration $T_{\mathrm{tr}}$, together with
\eqref{eq:planner_admissible_set}, prevents abrupt or excessive altitude
changes while preserving the nominal horizontal mission.

The intervals $[t_j^-,t_j^+]$ locate the vertical trajectory modifications
along the nominal mission timeline.
During closed-loop candidate propagation, however, disturbance exposure is
evaluated from the vehicle position using the spatial disturbance model in
Section~\ref{subsec:disturbance_model}.
The basis functions therefore locate the altitude modifications rather than
directly prescribing the disturbance experienced by the vehicle.

%%%%%%%%%%%%%%%%%%%%%%%%%%%%%%%%%%%%%%%%%%%%%%%%%%%%%%%%%%%%%%%%%%%%%%%%%%%
\subsection{Closed-Loop Battery-Aware Prediction}
\label{subsec:closed_loop_prediction}

Candidate trajectories are evaluated through closed-loop propagation rather
than solely from geometric or kinematic quantities.
For each $\boldsymbol{\delta}$, the reference
$\{\mathbf{p}_d,\dot{\mathbf{p}}_d,\ddot{\mathbf{p}}_d\}$ is propagated
through the vehicle model over the complete mission horizon.
At prediction step $k$, the feedback controller generates
\begin{equation}
	\mathbf{u}_k
	=
	\boldsymbol{\kappa}
	\left(
	\mathbf{x}_k,
	\mathbf{p}_{d,k},
	\dot{\mathbf{p}}_{d,k},
	\ddot{\mathbf{p}}_{d,k}
	\right),
	\label{eq:closed_loop_control_prediction}
\end{equation}
where $\boldsymbol{\kappa}(\cdot)$ denotes the feedback-control law.
The controller used during planning is held fixed throughout the optimization.
Control allocation then determines
$\boldsymbol{\omega}_{r,\mathrm{req},k}$ and
$\bar{\boldsymbol{\omega}}_{r,\mathrm{req},k}$, while the battery--actuator
coupling determines the achievable rotor commands and electrical load.

At each prediction step, terminal voltage, battery-dependent rotor-speed
capability, rotor command, electrical power, and battery current are mutually
coupled.
For fixed differential states
$\{s,V_{p,1},V_{p,2},\boldsymbol{\omega}_r\}$, the algebraic iteration is
initialized as
\begin{equation}
	V_b^{(0)}
	=
	\max
	\left\{
	2.8N_s,\,
	V_{\mathrm{oc}}(s)-V_{p,1}-V_{p,2}
	\right\}.
	\label{eq:fixed_point_initialization}
\end{equation}
The $2.8N_s$-V lower bound is used only as a numerical continuation safeguard
for the algebraic battery--actuator iteration and is distinct from the
mission-feasibility threshold $V_{\min}$.
Defining
\begin{equation}
	\bar V_b
	=
	V_{\mathrm{oc}}(s)-V_{p,1}-V_{p,2},
	\label{eq:planner_available_voltage}
\end{equation}
the coupled quantities are iteratively updated according to
\begin{subequations}
	\label{eq:fixed_point_prediction}
	\begin{align}
		\omega_{\max}^{(\ell)}
		&=
		\omega_{\max}\!\left(V_b^{(\ell)}\right),
		\\
		\boldsymbol{\omega}_{r,\mathrm{cmd}}^{(\ell)}
		&=
		\min
		\left\{
		\bar{\boldsymbol{\omega}}_{r,\mathrm{req}},
		\omega_{\max}^{(\ell)}\mathbf{1}
		\right\},
		\\
		P_b^{(\ell)}
		&=
		P_b\!\left(
		\boldsymbol{\omega}_r,
		\boldsymbol{\omega}_{r,\mathrm{cmd}}^{(\ell)}
		\right),
		\\
		I_b^{(\ell)}
		&=
		\frac{
			\bar V_b-
			\sqrt{
				\bar V_b^{\,2}-4R_0(s)P_b^{(\ell)}
			}
		}{
			2R_0(s)
		},
		\\
		\widetilde V_b^{(\ell+1)}
		&=
		\bar V_b-R_0(s)I_b^{(\ell)},
		\\
		V_b^{(\ell+1)}
		&=
		\lambda\widetilde V_b^{(\ell+1)}
		+
		(1-\lambda)V_b^{(\ell)},
	\end{align}
\end{subequations}
where the minimum in the rotor-command update is applied elementwise and
$\lambda\in(0,1]$ is a relaxation factor.
The iteration terminates when
\begin{equation}
	\left|
	V_b^{(\ell+1)}-V_b^{(\ell)}
	\right|
	\leq\varepsilon_V
	\label{eq:fixed_point_convergence}
\end{equation}
or when the prescribed maximum number of iterations $\ell_{\max}$ is reached.
Failure to satisfy \eqref{eq:fixed_point_convergence} within
$\ell_{\max}$ iterations is recorded as a coupling-convergence violation.
In the numerical study,
$\lambda=0.60$, $\varepsilon_V=10^{-5}$~V, and
$\ell_{\max}=30$.
The battery differential states remain fixed during this algebraic iteration
and are propagated only after the coupling has been resolved.
A real-valued discharge-current solution requires
\begin{equation}
	\bar V_b^{\,2}
	-
	4R_0(s)P_b
	\geq0.
	\label{eq:planner_power_discriminant}
\end{equation}
If the condition in \eqref{eq:planner_power_discriminant} is violated, the
candidate is flagged as electrically infeasible.
Because the quadratic battery-current relation no longer admits a real-valued
solution, a finite numerical continuation is used solely to maintain
candidate propagation and objective evaluation.
A bounded surrogate evaluation of current and terminal voltage is used solely
for numerical continuation, with terminal voltage lower-bounded by
$2.8N_s$~V.
The electrical-feasibility violation is retained in the planning penalty and
the final mission-feasibility check.
The corresponding instantaneous battery-power utilization is
\begin{eqnarray}
	 \eta_P(t) &=& \frac{P_b(t)}{P_{b,\max}(t)}, \\
 	P_{b,\max}(t) &=& \frac{
 		\left[
 		V_{\mathrm{oc}}(s(t))
 		-
 		V_{p,1}(t)
 		-
 		V_{p,2}(t)
 		\right]^2
 	}{
 		4R_0(s(t))
 	}.
 	\label{eq:battery_power_utilization}
\end{eqnarray}
 
Thus, $\eta_P\leq1$ corresponds to the instantaneous power-delivery
condition of the equivalent-circuit model. 
After the algebraic coupling is resolved, the resulting current propagates
$s$, $V_{p,1}$, and $V_{p,2}$ through the battery dynamics, while the
converged rotor command propagates the motor states.
Repeating this procedure over the complete mission provides the quantities
required to evaluate each candidate trajectory.

%%%%%%%%%%%%%%%%%%%%%%%%%%%%%%%%%%%%%%%%%%%%%%%%%%%%%%%%%%%%%%%%%%%%%%%%%%%
\subsection{Objective Function and Mission Feasibility}
\label{subsec:planner_objective}

The planner employs a normalized composite objective that balances electrical
energy, trajectory modification, and closed-loop tracking performance while
penalizing undesirable battery, actuator, and terminal operating conditions:
\begin{equation}
	\begin{split}
		J(\boldsymbol{\delta})={}&
		w_EJ_E+w_pJ_p+w_eJ_e+w_fJ_f
		+w_sJ_s+w_VJ_V\\
		&+w_{\Delta\omega}J_{\Delta\omega}
		+w_{\eta}J_{\eta}
		+w_{\mathrm{end}}J_{\mathrm{end}}
		+w_{\mathrm{phy}}J_{\mathrm{phy}},
	\end{split}
	\label{eq:planner_cost}
\end{equation}
where the $w_{(\cdot)}$ are nonnegative weighting coefficients.

The closed-loop position RMSE and terminal position error are
\begin{subequations}
	\label{eq:tracking_metrics}
	\begin{align}
		e_{\mathrm{RMSE}}
		&=
		\sqrt{
			\frac{1}{T_f}
			\int_{0}^{T_f}
			\left\|
			\mathbf{p}(t)-\mathbf{p}_d(t)
			\right\|_2^2\,dt
		},
		\\
		e_f
		&=
		\left\|
		\mathbf{p}(T_f)-\mathbf{p}_d(T_f)
		\right\|_2.
	\end{align}
\end{subequations}
The nominal objective terms are
\begin{subequations}
	\label{eq:nominal_cost_terms}
	\begin{align}
		J_E
		&=
		\frac{E_b(T_f)}{E_{\mathrm{nom}}},
		&
		J_p
		&=
		\frac{1}{N_d}
		\sum_{j=1}^{N_d}
		\left(
		\frac{\delta_j}{\delta_{\max}}
		\right)^2,
		\\
		J_e
		&=
		\left(
		\frac{e_{\mathrm{RMSE}}}{L_m}
		\right)^2,
		&
		J_f
		&=
		\left(
		\frac{e_f}{L_m}
		\right)^2,
	\end{align}
\end{subequations}
where $E_{\mathrm{nom}}$ is the nominal battery-energy scale and $L_m$ is a
characteristic mission-distance scale.

Battery, actuator, and endpoint requirements are incorporated using one-sided
quadratic penalties.
Defining $[x]_+=\max\{x,0\}$,
\begin{subequations}
	\label{eq:constraint_penalties}
	\begin{align}
		J_s
		&=
		\left(
		\frac{
			[s_{\min}-\min_t s(t)]_+
		}{
			s_{\min}
		}
		\right)^2,
		\\
		J_V
		&=
		\left(
		\frac{
			[V_{\min}-\min_tV_b(t)]_+
		}{
			V_{\min}
		}
		\right)^2,
		\\
		J_{\Delta\omega}
		&=
		\left(
		\frac{
			[\Delta\omega_{\min}
			-\min_t\Delta\omega(t)]_+
		}{
			\omega_{\max}^{\mathrm{nom}}
		}
		\right)^2,
		\\
		J_{\eta}
		&=
		\left[
		\max_t\eta_{\omega}(t)-\eta_{\max}
		\right]_+^2,
		\\
		J_{\mathrm{end}}
		&=
		\left(
		\frac{
			[e_f-e_{\max}]_+
		}{
			L_m
		}
		\right)^2.
	\end{align}
\end{subequations}
Here, $J_f$ continuously penalizes terminal tracking error, whereas
$J_{\mathrm{end}}$ becomes active only when the prescribed endpoint
threshold is exceeded.
Similarly, $J_{\Delta\omega}$ encourages a prescribed actuator reserve and
$J_{\eta}$ discourages operation above the preferred battery-dependent
utilization threshold.
These preferred thresholds shape the planning objective but do not define the
physical actuator boundary.

To reject candidates that violate the physical battery or actuator model, the
implementation additionally defines
\begin{equation}
	\begin{split}
		J_{\mathrm{phy}}
		={}&
		[\max_t\eta_P(t)-1]_+^2
		+
		[\max_t\eta_{\omega,\mathrm{nom}}(t)-1]_+^2
		\\
		&+
		\frac{1}{N_p}
		\sum_{k=1}^{N_p}\mathbb{I}_{\mathrm{elec},k}
		+
		\frac{1}{N_p}
		\sum_{k=1}^{N_p}\mathbb{I}_{\mathrm{conv},k},
	\end{split}
	\label{eq:physical_infeasibility_penalty}
\end{equation}
where
\begin{equation}
	\eta_{\omega,\mathrm{nom}}(t)
	=
	\frac{
		\max_i\omega_{i,\mathrm{req}}(t)
	}{
		\omega_{\max}^{\mathrm{nom}}
	},
	\label{eq:nominal_authority_utilization}
\end{equation}
$\mathbb{I}_{\mathrm{elec},k}=1$ when the battery power-delivery condition is
violated, $\mathbb{I}_{\mathrm{conv},k}=1$ when the battery--actuator
iteration does not converge, and $N_p$ is the number of prediction samples.
Thus, physically inadmissible candidates remain numerically evaluable but incur an explicit physical-infeasibility penalty during optimization.

The trajectory conditions in \eqref{eq:planner_admissible_set} are enforced
as explicit nonlinear constraints.
In contrast, SOC, terminal voltage, preferred actuator reserve, preferred
utilization, endpoint accuracy, and physical infeasibility enter the
optimization through the penalty terms above.
Following optimization, the selected trajectory is subjected to explicit
mission-feasibility checks.
A mission is classified as feasible only if
\begin{equation}
	\begin{aligned}
		\min_t s(t)
		&\geq s_{\min},
		&
		\min_tV_b(t)
		&\geq V_{\min},
		\\
		\max_t\eta_P(t)
		&\leq1,
		&
		\max_t\eta_{\omega,\mathrm{nom}}(t)
		&\leq1,
		\\
		\max_t\eta_\omega(t)
		&\leq1,
		&
		e_f
		&\leq e_{\max},
	\end{aligned}
	\label{eq:mission_feasibility}
\end{equation}
with successful battery--actuator coupling required throughout the mission.
The nominal utilization condition verifies the raw rotor-speed request
against the nominal hardware limit, whereas $\eta_\omega\leq1$ verifies the
nominally clipped request against the instantaneous battery-dependent
capability.

%%%%%%%%%%%%%%%%%%%%%%%%%%%%%%%%%%%%%%%%%%%%%%%%%%%%%%%%%%%%%%%%%%%%%%%%%%%
%\subsection{Numerical Optimization}
%\label{subsec:planner_numerical_optimization}
%
%The resulting optimization is nonlinear and generally piecewise smooth
%because candidate evaluation includes closed-loop simulation, one-sided
%penalties, saturation, and extrema over the mission horizon.
%%
%The problem is solved using sequential quadratic programming (SQP).
%%
%To reduce sensitivity to initialization, the optimization is repeated from
%multiple initial altitude-offset vectors, and the converged candidate having
%the lowest predicted objective is retained.
%%
%The numerical study uses the three initializations
%%
%\begin{equation}
%	\boldsymbol{\delta}^{(0)}
%	\in
%	\left\{
%	\begin{bmatrix}0&0&0\end{bmatrix}^{\top},
%	\begin{bmatrix}2.5&1.5&3.0\end{bmatrix}^{\top},  
%	\begin{bmatrix}-2&-1&-2.5\end{bmatrix}^{\top}
%	\right\}\mathrm{m}.
%	\label{eq:planner_initializations}
%\end{equation}
%%
%The prediction step is $\Delta t_p=0.05$~s.
%%
%The SQP solver uses a maximum of $45$ iterations and $180$ function
%evaluations, with step, optimality, and constraint tolerances of
%$10^{-3}$, $10^{-5}$, and $10^{-6}$, respectively.
%%
%The battery--actuator algebraic iteration uses a maximum of $30$ iterations,
%a terminal-voltage tolerance of $10^{-5}$~V, and a relaxation factor
%$\lambda=0.60$.
%%
%The resulting optimizer convergence and computational performance are
%reported with the numerical results.
%%%%%%%%%%%%%%%%%%%%%%%%%%%%%%%%%%%%%%%%%%%%%%%%%%%%%%%%%%%%%%%%%%%%%%%%%%%%%%%%
 \section{Simulation Results and Discussion}
 \label{sec:simulation}
 
The proposed predictive trajectory-planning and control framework is evaluated in a long-duration, high-speed multirotor mission containing three spatially localized disturbance regions.
 The simulations examine the interaction among trajectory selection,
 disturbance exposure, closed-loop tracking, electrical energy consumption,
 battery evolution, and battery-dependent actuator capability.
The reduced-order battery model is first numerically benchmarked against an
independently implemented Simscape equivalent-circuit reference.
 The nominal and battery-aware trajectories are then compared over the
 complete mission, followed by planner ablations at nominal and reduced
 initial SOC.
 Finally, the selected battery-aware trajectory is executed using multiple
 feedback controllers to examine the sensitivity of mission performance to
 the closed-loop control architecture.

 \subsection{Simulation Setup and Mission Description}
 \label{subsec:simulation_setup}
 
 The simulated multirotor has a mass of $m=1.8$~kg, arm length $l=0.23$~m,
 and inertia matrix
 \begin{equation}
 	\mathbf{J}
 	=
 	\operatorname{diag}
 	\left(
 	0.030,\,
 	0.030,\,
 	0.055
 	\right)
 	\ \mathrm{kg\,m^2}.
 	\label{eq:simulation_inertia}
 \end{equation}
 The thrust and reaction-torque coefficients are
 $k_T=1.05\times10^{-5}$ and $k_Q=1.7\times10^{-7}$, respectively.
 The motor time constant is $\tau_m=0.035$~s, the nominal maximum rotor
 speed is $\omega_{\max}^{\mathrm{nom}}=1050$~rad/s, and the propulsion
 efficiency is $\eta_m=0.82$.
 The vehicle is powered by a four-cell lithium-ion battery with a rated
 capacity of $Q_b=5$~Ah.
 Unless otherwise stated, the initial SOC is $92\%$.
 %
% The SOC-dependent reduced-order battery model described in Section~III is
% used during both predictive trajectory evaluation and closed-loop mission
% simulation.
% %
% An auxiliary electrical load of $P_{\mathrm{aux}}=8$~W is included in
% addition to the propulsion load.
 %
 The closed-loop vehicle simulation uses a time step of
 $\Delta t=0.005$~s, whereas candidate trajectories are evaluated by the
 planner using $\Delta t_p=0.05$~s.
 The $150$-s mission consists of takeoff over $0$--$10$~s, horizontal
 flight over $10$--$140$~s, and landing over $140$--$150$~s.
 The nominal flight altitude is $5$~m.
 During horizontal flight, a quintic reference is generated through the prescribed waypoints, with zero acceleration specified at the segment boundaries.
 \begin{equation}
 	\begin{aligned}
 		\mathbf{t}_w
 		&= [10,\;30,\;70,\;110,\;140]~\mathrm{s},\\
 		\mathbf{x}_w
 		&= [0,\;60,\;220,\;460,\;640]~\mathrm{m},\\
 		\dot{\mathbf{x}}_w
 		&= [0,\;3,\;5,\;6,\;0]~\mathrm{m/s},
 	\end{aligned}
 	\label{eq:mission_waypoints}
 \end{equation}
 The resulting $640$-m mission exposes the vehicle to disturbances at different operating speeds and battery states.
 \begin{table}[h]
	\centering
	\caption{Disturbance conditions for the long-duration mission.}
	\label{tab:disturbance_setup}
	\begin{tblr}{
			width = \columnwidth,
			colspec = {X[0.55,c] X[0.85,c] X[0.95,c] X[0.95,c]},
			hlines = {1pt},
			vlines = {0.6pt},
		}
		Disturbance & Nominal encounter (s) &
		$v_{w,y}$ (m/s) & $v_{w,z}$ (m/s)\\
		$D_1$ & $30$--$42$   & $12$ & $-3.75$\\
		$D_2$ & $72$--$84$   & $7$  & $-2.19$\\
		$D_3$ & $112$--$124$ & $14$ & $-4.38$\\
	\end{tblr}
\end{table}
 
 Three spatially localized disturbance regions are introduced during the
 horizontal portion of the mission, as summarized in
 Table~\ref{tab:disturbance_setup}.
 The vertical wind component is defined as
 $v_{w,z}=-0.3125v_{w,y}$.
 The disturbance field is centered at the nominal flight altitude of $5$~m
 with a vertical spread of $\sigma_d=1.25$~m.
 
The planner comparison considers three information and objective structures.
The \emph{disturbance-aware} case executes the nominal reference trajectory
without geometric modification while accounting for the prescribed
disturbance field during closed-loop propagation.
The \emph{energy-aware, battery-unaware} planner modifies the reference
trajectory based on the predicted closed-loop electrical-energy consequences
of disturbance exposure, but excludes battery-state-dependent voltage and
actuator information from the trajectory-selection process.
Finally, the proposed \emph{battery-aware} planner evaluates candidate
trajectories through the complete closed-loop vehicle--motor--battery
prediction and additionally accounts for the battery-dependent quantities
defined in Section~\ref{sec:predictive_planner}.
All selected trajectories are subsequently evaluated using the same coupled
vehicle--motor--battery plant.
This comparison isolates the benefit of energy-aware trajectory modification
from the additional influence of battery-dependent feasibility information.
 
The nonlinear trajectory optimization is solved using SQP from the three initial altitude-offset vectors
\begin{equation}
	\boldsymbol{\delta}^{(0)}
	\in
	\left\{
	\begin{aligned}
		&\begin{bmatrix}0&0&0\end{bmatrix}^{\top},
		\quad
		\begin{bmatrix}2.5&1.5&3.0\end{bmatrix}^{\top},\\
		&\begin{bmatrix}-2&-1&-2.5\end{bmatrix}^{\top}
	\end{aligned}
	\right\}\mathrm{m}.
	\label{eq:planner_initializations_results}
\end{equation}
 Because the resulting nonlinear optimization problem is nonconvex, the
 reported trajectory is the lowest-cost solution obtained from the tested
 initializations and is not claimed to be globally optimal.

 \subsection{Battery Model Numerical Verification}
 \label{subsec:battery_validation}

The explicit battery model used for candidate-trajectory propagation is
numerically benchmarked against an independently implemented Simscape
equivalent-circuit reference using the same SOC-dependent parameterization.
The comparison therefore evaluates agreement between the
algebraic--differential implementation used by the planner and the Simscape
reference implementation; it is not intended as experimental battery-model
validation.
Both models are driven by the same mission-representative electrical-power
profile and initialized from identical battery conditions.
 
 \begin{figure}[h]
 	\centering
 	\includegraphics[width=\linewidth]{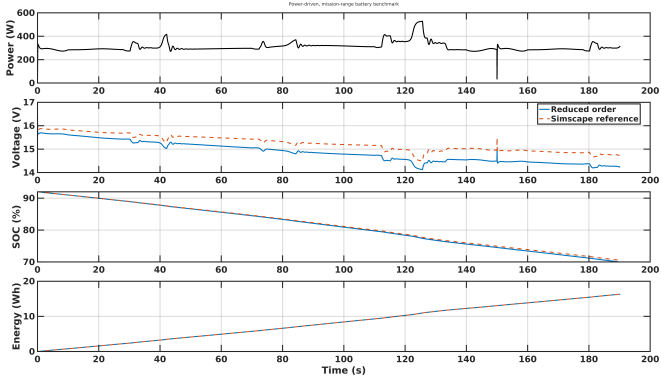}
 	\caption{Numerical verification of the battery model used for
 		candidate-trajectory propagation against the independently implemented
 		Simscape equivalent-circuit reference under the same prescribed
 		electrical-power profile and SOC-dependent parameterization.}
 	\label{fig:battery_validation}
 \end{figure}
 
  \begin{figure*}[!ht]
 	\centering
 	\includegraphics[width=0.97\textwidth]{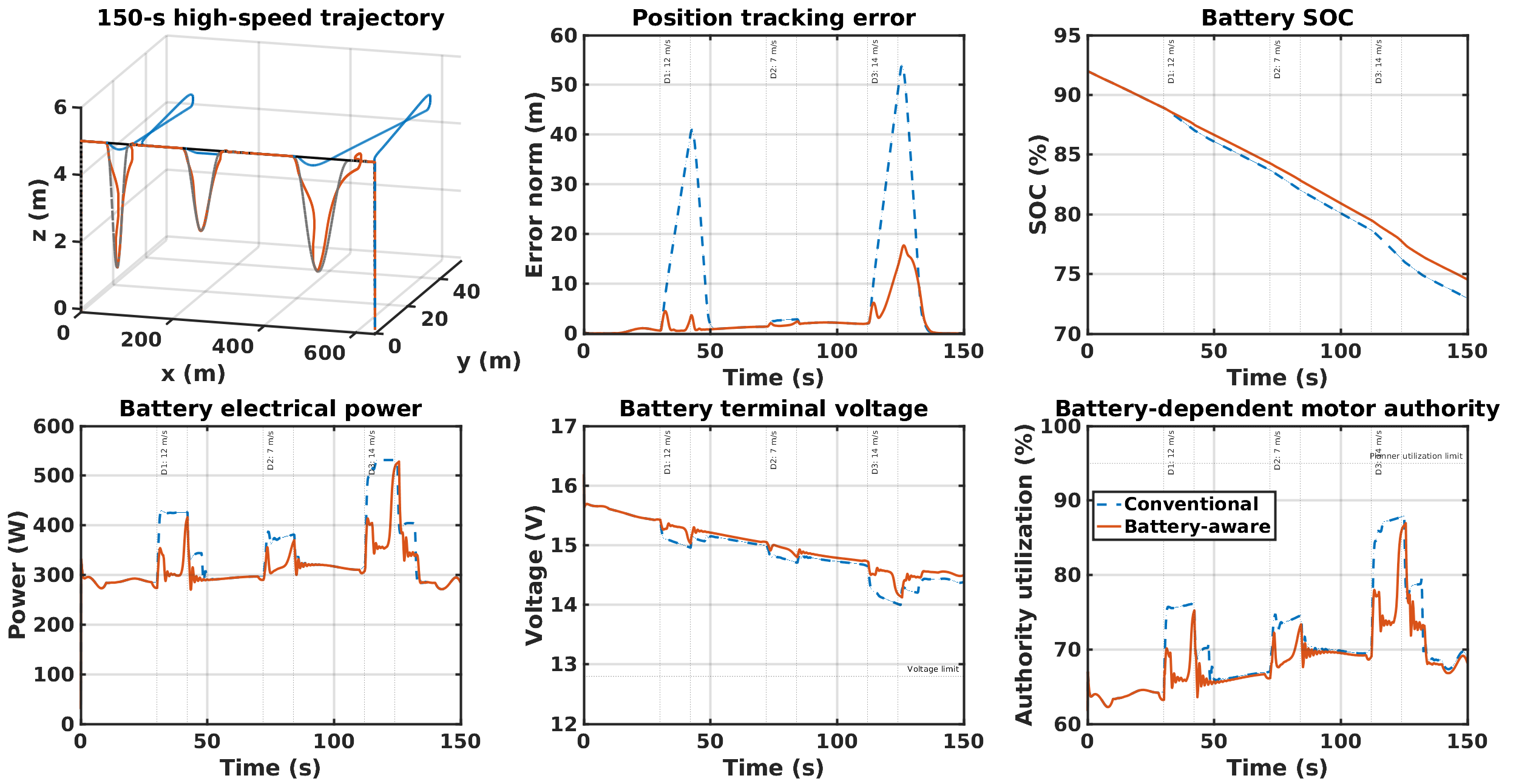}
 	\caption{Closed-loop mission response for the conventional and
 		battery-aware trajectories over the $150$-s mission. The panels show
 		the realized trajectory, position-tracking error, battery SOC,
 		electrical power, terminal voltage, and battery-dependent actuator
 		utilization. Vertical markers identify the nominal disturbance
 		encounters.}
 	\label{fig:comparison_dashboard}
 \end{figure*}
 As shown in Fig.~\ref{fig:battery_validation}, the reduced-order model
 closely reproduces the SOC and cumulative-energy evolution of the Simscape
 reference over the complete discharge profile.
 The SOC RMSE is $0.0239$ percentage points, while the terminal-voltage RMSE
 is $0.1095$~V.
 The power RMSE is $2.2205$~W, corresponding to a normalized RMSE of
 $0.6379\%$ over the reference-model power range.
 The cumulative energies predicted by the reduced-order and Simscape models
 are $2.1155$ and $2.1010$~Wh, respectively, corresponding to a discrepancy
 of $0.6852\%$.
 The voltage comparison exhibits a larger systematic discrepancy than the SOC
 and cumulative-energy responses, but preserves the load-dependent voltage
 depressions required for the subsequent actuator-capability calculation.
 These results provide a numerical model-to-model benchmark supporting use
 of the reduced-order model for repeated planner evaluation; they are not
 presented as experimental battery validation.

 \subsection{Closed-Loop Mission Response}
 \label{subsec:closed_loop_results}
  
 Fig.~\ref{fig:comparison_dashboard} compares the conventional and
 battery-aware closed-loop responses over the complete $150$-s mission.
 The conventional reference remains near the center of the disturbance
 layer, whereas the battery-aware trajectory introduces altitude deviations during the three disturbance encounters.
 For the nominal $92\%$ initial-SOC condition, the selected battery-aware
 offsets are
 \begin{equation}
 	\boldsymbol{\delta}^{*}
 	=
 	\begin{bmatrix}
 		-3.7076 & -2.4226 & -3.3912
 	\end{bmatrix}^{\top}\mathrm{m}.
 	\label{eq:optimized_offsets_results}
 \end{equation}
 Thus, the selected reference moves below the disturbance center during all
 three encounters.
  
 The conventional trajectory exhibits large tracking excursions during the
 first and third disturbance encounters, with the largest error occurring
 during $D_3$.
 The battery-aware trajectory substantially reduces these excursions by
 moving away from the center of the spatial disturbance field.
 The mission-level position RMSE decreases from approximately $15.52$~m to
 $4.34$~m, corresponding to a reduction of approximately $72\%$.
 
 The improved tracking is accompanied by lower cumulative electrical demand.
 The conventional trajectory consumes approximately $14.11$~Wh, whereas the
 battery-aware trajectory consumes approximately $13.06$~Wh, corresponding
 to an energy reduction of approximately $7.5\%$.
 The reduced electrical demand results in a higher retained SOC and terminal
 voltage during the later portion of the mission.
 
 The power response also shows that the energetic improvement does not result
 from reducing instantaneous power at every time instant.
 The altitude-transition maneuvers introduce short-duration increases in
 control and electrical demand.
 These increases are offset by the reduced feedback effort required while the
 vehicle traverses the disturbance regions.
 The trajectory modification therefore redistributes control effort over the
 mission such that the cumulative energetic consequence of disturbance
 exposure is reduced.
 
 The actuator response provides an important qualification to these results.
 At the nominal initial SOC, both trajectories remain below the
 battery-dependent actuator limit, with peak utilization remaining below
 approximately $90\%$.
 The battery-dependent actuator constraint is therefore nonbinding in this
 case.
\begin{table*}[!ht]
	\centering
	\caption{Planning-ablation results for the nominal $92\%$ initial SOC.}
	\label{tab:planner_ablation_results}
	\begin{tblr}{
			width = \linewidth,
			colspec = {
				X[1.35,l,m]
				X[0.6,c,m]
				X[0.6,c,m]
				X[0.6,c,m]
				X[0.78,c,m]
				X[0.82,c,m]
				X[0.78,c,m]
				X[0.8,c,m]
			},
			hlines = {1pt},
			vlines = {0.6pt},
		}
		Case &
		$\delta_1$ (m) &
		$\delta_2$ (m) &
		$\delta_3$ (m) &
		Energy (Wh) &
		Energy red. (\%) &
		RMSE (m) &
		RMSE red. (\%)\\
		
		Disturbance-aware
		& 0 & 0 & 0
		& 14.1078 & 0
		& 15.5226 & 0\\
		
		Energy-aware, battery-unaware
		& -3.7076 & -2.4226 & -3.3912
		& 13.0552 & 7.46
		& 4.3461 & 72.0\\
		
		Battery-aware
		& -3.7076 & -2.4226 & -3.3912
		& 13.0552 & 7.46
		& 4.3461 & 72.0\\
	\end{tblr}
\end{table*}
\begin{table*}[!hb]
	\centering
	\caption{Planning-ablation results for the low-SOC stress condition
		($SOC(0)=55\%$).}
	\label{tab:stress_results}
	\begin{tblr}{
			width = \linewidth,
			colspec = {
				X[1.35,l,m]
				X[0.58,c,m]
				X[0.58,c,m]
				X[0.58,c,m]
				X[0.75,c,m]
				X[0.75,c,m]
				X[0.75,c,m]
				X[0.82,c,m]
				X[0.62,c,m]
			},
			hlines = {1pt},
			vlines = {0.6pt},
		}
		Case &
		$\delta_1$ (m) &
		$\delta_2$ (m) &
		$\delta_3$ (m) &
		Energy (Wh) &
		RMSE (m) &
		$V_{b,\min}$ (V) &
		$\eta_{\omega,\max}$ (\%) &
		Feasible\\
		
		Disturbance-aware
		& 0 & 0 & 0
		& 15.0825
		& 89.9623
		& 11.6457
		& 137.3898
		& No\\
		
		Energy-aware, battery-unaware
		& -3.7075 & -2.4226 & -3.3913
		& 13.0494
		& 4.3385
		& 12.0732
		& 107.6531
		& No\\
		
		Battery-aware
		& -4.9570 & -3.6519 & -4.0461
		& 12.9801
		& 4.3048
		& 12.0814
		& 107.4351
		& No\\
	\end{tblr}
\end{table*}
Accordingly, the nominal-SOC improvement is primarily attributable to
reduced disturbance exposure and the associated decrease in closed-loop
control and electrical demand rather than to avoidance of an active
battery-dependent actuator constraint.
\subsection{Planning Ablation}
\label{subsec:planner_ablation}

Table~\ref{tab:planner_ablation_results} summarizes the planning-ablation
results for the nominal initial SOC of $92\%$.
Both trajectory-modifying planners select
\begin{equation}
	\boldsymbol{\delta}^{*}_{E}
	=
	\boldsymbol{\delta}^{*}_{B}
	=
	\begin{bmatrix}
		-3.7076 & -2.4226 & -3.3912
	\end{bmatrix}^{\top}\mathrm{m}.
	\label{eq:nominal_soc_offsets}
\end{equation}
Relative to the disturbance-aware baseline, the resulting trajectory reduces
mission energy from $14.1078$ to $13.0552$~Wh, corresponding to a $7.46\%$
reduction, while decreasing the position-tracking RMSE from $15.5226$ to
$4.3461$~m, corresponding to a $72.0\%$ reduction.
The offsets move the reference below the nominal altitude during the
three disturbance encounters, reducing the closed-loop control demand
associated with disturbance exposure.

The equality of the energy-aware battery-unaware and battery-aware solutions
at $92\%$ initial SOC is itself an important ablation result.
Under this condition, the battery-dependent voltage and actuator terms remain
nonbinding along the selected trajectory and therefore do not alter the
lowest-cost solution obtained from the tested initializations.
The nominal-SOC results consequently attribute the observed energy and
tracking improvements primarily to trajectory modification based on predicted
closed-loop energetic consequences rather than to avoidance of an active
battery-dependent constraint.

To determine whether battery-dependent information affects the planning
decision as the available electrical and actuator capability decreases, the
same comparison is repeated with the initial SOC reduced to $55\%$.
All other mission, vehicle, disturbance, controller, and planning parameters
are unchanged.
The resulting stress-test outcomes are summarized in
Table~\ref{tab:stress_results}.
 
The energy-aware battery-unaware planner selects
\begin{equation}
	\boldsymbol{\delta}^{*}_{E,\mathrm{stress}}
	=
	\begin{bmatrix}
		-3.7075 & -2.4226 & -3.3913
	\end{bmatrix}^{\top}\mathrm{m},
	\label{eq:stress_energy_offsets}
\end{equation}
which is essentially unchanged from its $92\%$-SOC solution.
This behavior is expected because battery-state-dependent voltage and
actuator information does not enter the energy-aware planner's objective.
In contrast, the battery-aware planner selects
\begin{equation}
	\boldsymbol{\delta}^{*}_{B,\mathrm{stress}}
	=
	\begin{bmatrix}
		-4.9570 & -3.6519 & -4.0461
	\end{bmatrix}^{\top}\mathrm{m},
	\label{eq:stress_battery_offsets}
\end{equation}
with larger-magnitude altitude deviations during all three disturbance
encounters.
The separation between
$\boldsymbol{\delta}^{*}_{E,\mathrm{stress}}$ and
$\boldsymbol{\delta}^{*}_{B,\mathrm{stress}}$, in contrast to their equality
at $92\%$ SOC, demonstrates that the battery-dependent terms have become
consequential to the planning decision.

The modified battery-aware trajectory also produces modest improvements
relative to the energy-aware battery-unaware trajectory under the stress
condition.
Mission energy decreases from $13.0494$ to $12.9801$~Wh, while the
position-tracking RMSE decreases from $4.3385$ to $4.3048$~m.
The minimum terminal voltage increases from $12.0732$ to $12.0814$~V, and
the maximum battery-dependent actuator utilization decreases from
$107.6531\%$ to $107.4351\%$.
Although these changes are small in the aggregate mission metrics, the
different selected offsets show that battery-state-dependent information
changes the optimizer's preferred trajectory once the electrical and
actuator limits become restrictive.
\begin{figure*}[!ht] 
	\centering
	\includegraphics[width=\linewidth]{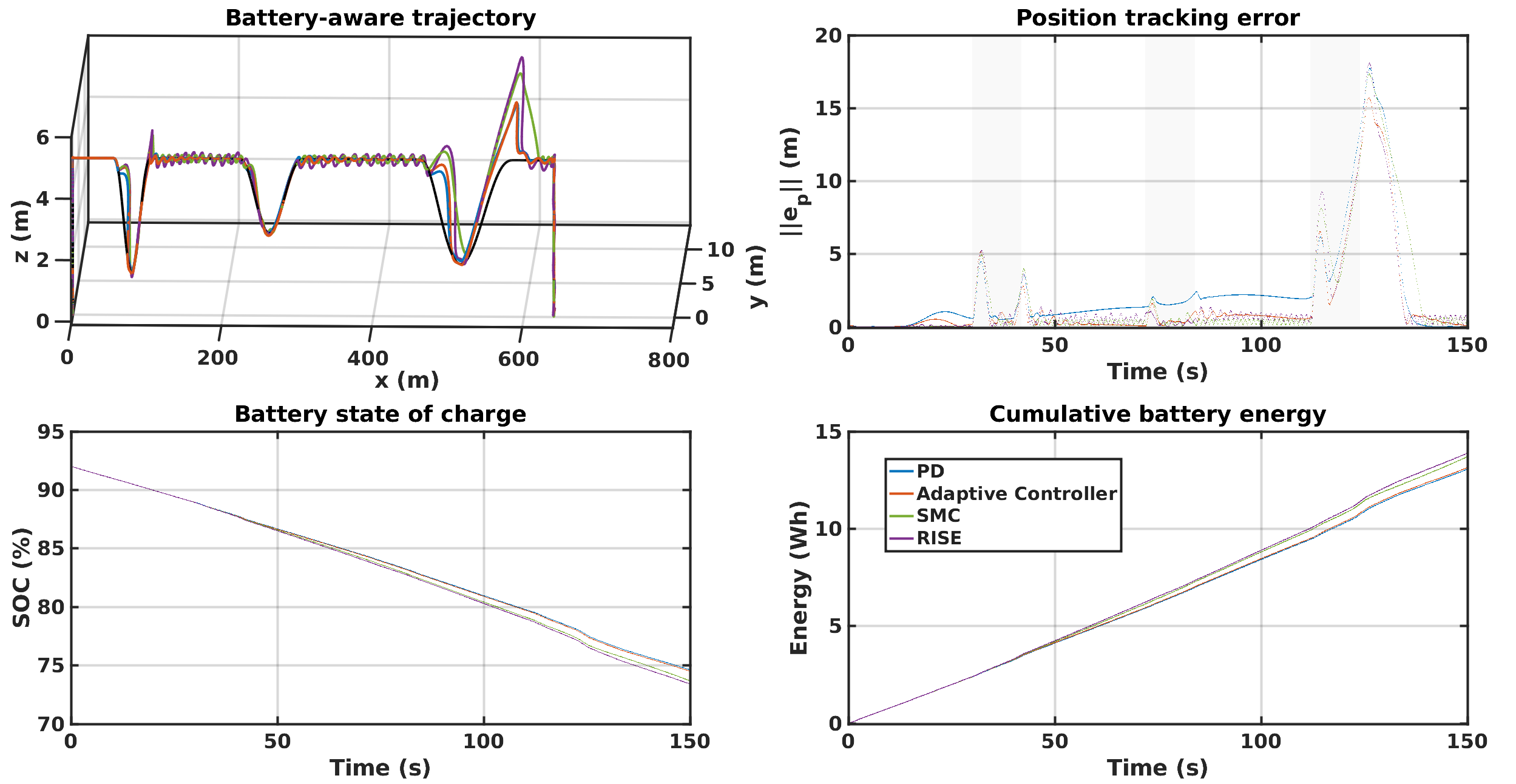} 
	\caption{Execution of the fixed battery-aware trajectory using PD, adaptive, SMC, and RISE controllers. The panels compare the realized trajectory, position-tracking error, battery SOC, and cumulative electrical energy.} \label{fig:controller_comparison} 
\end{figure*}
\begin{table*}[!hb]
	\centering
	\caption{Closed-loop performance of the battery-aware trajectory
		across different feedback controllers.}
	\label{tab:controller_comparison}
	\begin{tblr}{
			width = \linewidth,
			colspec = {X[1.0,l] *{6}{X[0.8,c]}},
			hlines = {1pt},
			vlines = {0.6pt},
		}
		Controller &
		RMSE (m) &
		Final error (m) &
		Energy (Wh) &
		$V_{b,\min}$ (V) &
		$\eta_{\omega,\max}$ (\%) &
		Feasible \\
		
		PD
		& 4.3407 & 0.0057 & 13.0552 & 14.2177 & 86.37 & Yes \\
		
		Adaptive
		& 3.6881 & 0.0875 & 13.1363 & 14.1690 & 87.79 & Yes \\
		
		SMC
		& 4.2326 & 0.4906 & 13.6990 & 14.0838 & 89.43 & Yes \\
		
		RISE
		& 3.9097 & 0.4291 & 13.8867 & 14.0587 & 90.62 & Yes \\
	\end{tblr}
\end{table*}

The $55\%$ initial-SOC case is intentionally severe, and none of the three
cases satisfies all mission-feasibility requirements.
In particular, even the battery-aware trajectory reaches a minimum terminal
voltage of $12.0814$~V, below the prescribed $12.8$-V limit, and a maximum
battery-dependent actuator utilization of $107.4351\%$, above the $100\%$
physical limit.
The stress test is therefore not presented as a demonstration of
mission-feasibility recovery.
Rather, it provides a causal ablation showing that battery-dependent
information, which is inactive at nominal SOC, changes the selected
trajectory as battery and actuator limitations become restrictive.

Taken together, the two operating conditions separate the roles of energetic
and battery-dependent prediction in the proposed framework.
At nominal SOC, predictive trajectory shaping reduces disturbance-induced
closed-loop demand and thereby improves both tracking and mission energy,
while the battery-dependent terms remain nonbinding.
Under depleted-battery conditions, those terms alter the selected trajectory,
but trajectory modification alone cannot guarantee feasibility when the
available battery-dependent actuator capability is insufficient for the
prescribed mission.
 
 \subsection{Sensitivity to Feedback-Control Architecture}
 \label{subsec:controller_comparison}
 
 The preceding planner studies use the baseline cascaded PD controller.
 Because candidate trajectories are evaluated through their predicted
 closed-loop consequences, an additional study examines whether mission
 performance changes when the selected reference is executed using different
 feedback-control architectures.
 The same battery-aware trajectory is executed using cascaded PD
 \cite{lopez-sanchezPIDControlQuadrotor2023}, adaptive
 \cite{liuAdaptivePredefinedTimePrescribed2026}, sliding-mode (SMC)
 \cite{alkomyEnergySavingIntegralSliding2026}, and RISE
 \cite{kidambiRobustNonlinearControlBased2021} controllers.
 All cases use identical vehicle, motor, battery, disturbance,
 actuator-limit, reference-trajectory, and initial-state conditions.
 The comparison therefore isolates the effect of feedback control after the
 planner-level reference has been fixed.
 
 Fig.~\ref{fig:controller_comparison} shows that the feedback architecture
 noticeably changes the closed-loop response to the same planned reference,
 particularly during the stronger disturbance encounters.
 As summarized in Table~\ref{tab:controller_comparison}, the adaptive
 controller provides the lowest mission-level position RMSE,
 $3.6881$~m, compared with $4.3407$~m for PD, $4.2326$~m for SMC, and
 $3.9097$~m for RISE.
 Relative to PD, the adaptive, SMC, and RISE controllers reduce the RMSE by
 approximately $15.0\%$, $2.5\%$, and $9.9\%$, respectively.
 
 The tracking improvement is accompanied by different electrical and actuator
 demands.
 The PD case consumes $13.0552$~Wh, compared with $13.1363$, $13.6990$,
 and $13.8867$~Wh for the adaptive, SMC, and RISE controllers,
 respectively.
 Relative to PD, these values correspond to energy increases of approximately
 $0.62\%$, $4.93\%$, and $6.37\%$.
 The maximum battery-dependent actuator utilization similarly increases from
 $86.37\%$ for PD to $87.79\%$, $89.43\%$, and $90.62\%$ for the adaptive,
 SMC, and RISE controllers, respectively.
 The corresponding minimum terminal voltages are $14.2177$, $14.1690$,
 $14.0838$, and $14.0587$~V.
  The adaptive controller provides the strongest tracking--energy compromise
 among the controllers considered here, reducing the RMSE by approximately
 $15\%$ relative to PD while increasing mission energy by only $0.62\%$.
 RISE provides the second-lowest RMSE but incurs a substantially larger
 energetic and actuator cost.
 The final position errors are $0.0057$, $0.0875$, $0.4906$, and $0.4291$~m for the PD, adaptive, SMC, and RISE controllers.
 
 All four cases satisfy the prescribed mission-feasibility criteria.
 Consequently, the controller providing the lowest mission-level tracking
 error does not simultaneously minimize energy consumption, actuator
 utilization, or terminal error.
 These results reinforce the distinction between trajectory planning and
 feedback control in the proposed framework.
 The planner determines the reference according to predicted disturbance,
 energetic, battery, and actuator consequences, whereas the feedback
 controller determines how that reference is realized during execution.
 Controller selection therefore modifies the tradeoff among tracking
 accuracy, electrical energy, actuator utilization, and terminal performance
 even when the planned trajectory is held fixed.
%%%%%%%%%%%%%%%%%%%%%%%%%%%%%%%%%%%%%%%%%%%%%%%%%%%%%%%%%%%%%%%%%%%%%%%%%%%%%%%%
\section{Conclusion}
\label{sec:conclusion}

This paper presented a battery-aware predictive trajectory-planning framework that couples multirotor dynamics, closed-loop control, motor response, battery evolution, and voltage-dependent actuator capability.
Candidate trajectories were evaluated by propagating their predicted
closed-loop consequences over the mission, allowing the planner to account
for the energetic and battery--actuator effects of disturbance exposure.
In the nominal-SOC mission, the selected trajectory reduced electrical
energy consumption by $7.46\%$ and position-tracking RMSE by approximately
$72\%$.
Planner ablations showed that these improvements primarily resulted from
reduced disturbance exposure and electrical demand, since the
battery-dependent terms remained nonbinding at nominal SOC.
Under a depleted-battery stress condition, however, the battery-dependent
terms changed the selected trajectory, demonstrating their increasing
influence as electrical and actuator limits became restrictive.
Finally, execution of the fixed battery-aware trajectory with multiple
feedback controllers demonstrated a controller-dependent tradeoff among
tracking accuracy, energy consumption, and actuator utilization.
These results show that closed-loop prediction of
trajectory--control--battery interactions provides a systematic basis for
selecting disturbance-aware multirotor trajectories while accounting for the
evolving electrical and actuator consequences of mission execution.
Future work will extend the present mission-level formulation toward
receding-horizon planning with online disturbance estimation and battery-state
updates.
Of particular interest is the development of feasibility-aware replanning
strategies that can modify the trajectory, mission timing, or terminal
requirements when battery-dependent actuator capability becomes insufficient
to complete the original mission.
Experimental flight validation will also be pursued to assess the coupled
vehicle--battery predictions under measured disturbances and hardware
limitations.
%%%%%%%%%%%%%%%%%%%%%%%%%%%%%%%%%%%%%%%%%%%%%%%%%%%%%%%%%%%%%%%%%%%%%%%%%
\section*{Acknowledgment}
ChatGPT 5.1 [OpenAI] was used  in the preparation to improve the clarity, grammar, and readability of the text.
%%%%%%%%%%%%%%%%%%%%%%%%%%%%%%%%%%%%%%%%%
\printbibliography

@article{alkomyEnergySavingIntegralSliding2026,
  title = {Energy-{{Saving Integral Sliding Mode Controller}} for a {{Quadrotor With}} a {{Slung Payload}}},
  author = {Alkomy, Hassan and Wang, Hao and Shan, Jinjun},
  year = 2026,
  journal = {IEEE Control Systems Letters},
  volume = {10},
  pages = {1243--1248},
  doi = {10.1109/LCSYS.2026.3706923},
  urldate = {2026-08-31}
}

@article{baekBatteryAwareOperationRange2019a,
  title = {Battery-{{Aware Operation Range Estimation}} for {{Terrestrial}} and {{Aerial Electric Vehicles}}},
  author = {Baek, Donkyu and Chen, Yukai and Bocca, Alberto and Bottaccioli, Lorenzo and Cataldo, Santa Di and Gatteschi, Valentina and Pagliari, Daniele Jahier and Patti, Edoardo and Urgese, Gianvito and Chang, Naehyuck and Macii, Alberto and Macii, Enrico and Montuschi, Paolo and Poncino, Massimo},
  year = 2019,
  month = jun,
  journal = {IEEE Transactions on Vehicular Technology},
  volume = {68},
  number = {6},
  pages = {5471--5482},
  doi = {10.1109/TVT.2019.2910452},
  urldate = {2026-08-27}
}

@article{benarfaMotionEnergyHealthAware2026a,
  title = {From {{Motion}} to {{Energy}}: {{Health-Aware Motion Planning}} for {{Enhanced Predictive Energy Management}} in {{Hybrid Fuel Cell}}/{{Battery Autonomous Mobile Robots}}},
  shorttitle = {From {{Motion}} to {{Energy}}},
  author = {Benarfa, Ghofrane and Mohammadpour, Mohammad and H{\'e}bert, Marie and Kelouwani, Sousso and Amamou, Ali},
  year = 2026,
  journal = {IEEE Vehicular Technology Magazine},
  pages = {2--11},
  doi = {10.1109/MVT.2026.3684541},
  urldate = {2026-08-27}
}

@article{chenOnlineStateCharge2019,
  title = {Online State of Charge Estimation of {{Li-ion}} Battery Based on an Improved Unscented {{Kalman}} Filter Approach},
  author = {Chen, Zewang and Yang, Liwen and Zhao, Xiaobing and Wang, Youren and He, Zhijia},
  year = 2019,
  month = jun,
  journal = {Applied Mathematical Modelling},
  volume = {70},
  pages = {532--544},
  doi = {10.1016/j.apm.2019.01.031},
  urldate = {2026-09-04},
  langid = {english}
}

@article{choiAdaptiveNeuroFuzzySliding2025,
  title = {Adaptive {{Neuro-Fuzzy Sliding Mode Tracking}} for {{Quadrotor UAVs}}},
  author = {Choi, Hyun Duck and Kim, Kwan Soo and Shi, Peng and Ahn, Choon Ki},
  year = 2025,
  journal = {IEEE Transactions on Automation Science and Engineering},
  volume = {22},
  pages = {16322--16333},
  doi = {10.1109/TASE.2025.3576292},
  urldate = {2026-08-31}
}

@article{conteDatadrivenLearningMethod2022,
  title = {A Data-Driven Learning Method for Online Prediction of Drone Battery Discharge},
  author = {Conte, C. and Rufino, G. and De Alteriis, G. and Bottino, V. and Accardo, D.},
  year = 2022,
  month = nov,
  journal = {Aerospace Science and Technology},
  volume = {130},
  pages = {107921},
  doi = {10.1016/j.ast.2022.107921},
  urldate = {2026-05-22},
  langid = {english}
}

@article{daiDataefficientModelingPower2024,
  title = {Data-Efficient Modeling for Power Consumption Estimation of Quadrotor Operations Using Ensemble Learning},
  author = {Dai, Wei and Zhang, Mingcheng and Low, Kin Huat},
  year = 2024,
  month = jan,
  journal = {Aerospace Science and Technology},
  volume = {144},
  pages = {108791},
  doi = {10.1016/j.ast.2023.108791},
  urldate = {2025-12-10},
  langid = {english}
}

@misc{gascheEnergyAwareSafe2025,
  title = {Energy {{Aware}} and {{Safe Path Planning}} for {{Unmanned Aircraft Systems}}},
  author = {Gasche, Sebastian and Kallies, Christian and Himmel, Andreas and Findeisen, Rolf},
  year = 2025,
  month = apr,
  number = {arXiv:2504.03271},
  eprint = {2504.03271},
  primaryclass = {eess.SY},
  publisher = {arXiv},
  doi = {10.48550/arXiv.2504.03271},
  urldate = {2026-05-22},
  archiveprefix = {arXiv}
}

@article{gongModelingPowerConsumptions2023,
  title = {Modeling {{Power Consumptions}} for {{Multirotor UAVs}}},
  author = {Gong, Hao and Huang, Baoqi and Jia, Bing and Dai, Hansu},
  year = 2023,
  month = dec,
  journal = {IEEE Transactions on Aerospace and Electronic Systems},
  volume = {59},
  number = {6},
  pages = {7409--7422},
  doi = {10.1109/TAES.2023.3288846},
  urldate = {2026-09-04}
}

@inproceedings{greiffQuadrotorMotionPlanning2023,
  title = {Quadrotor {{Motion Planning}} in {{Stochastic Wind Fields}}},
  booktitle = {2023 {{American Control Conference}} ({{ACC}})},
  author = {Greiff, Marcus and Vinod, Abraham and Nabi, Saleh and Di Cairano, Stefano},
  year = 2023,
  month = may,
  pages = {4619--4625},
  doi = {10.23919/ACC55779.2023.10155844},
  urldate = {2026-08-31}
}

@article{guLearningUncertaintiesOnline2025,
  title = {Learning Uncertainties Online for Quadrotor Flight Control: {{A}} Comparative Study},
  shorttitle = {Learning Uncertainties Online for Quadrotor Flight Control},
  author = {Gu, Weibin and Zhao, Jiance and Rizzo, Alessandro},
  year = 2025,
  month = sep,
  journal = {Journal of Intelligent \& Robotic Systems},
  volume = {111},
  number = {3},
  pages = {98},
  doi = {10.1007/s10846-025-02305-5},
  urldate = {2026-08-31},
  langid = {english}
}

@article{idrissiReviewQuadrotorUnmanned2022,
  title = {A {{Review}} of {{Quadrotor Unmanned Aerial Vehicles}}: {{Applications}}, {{Architectural Design}} and {{Control Algorithms}}},
  shorttitle = {A {{Review}} of {{Quadrotor Unmanned Aerial Vehicles}}},
  author = {Idrissi, Moad and Salami, Mohammad and Annaz, Fawaz},
  year = 2022,
  month = jan,
  journal = {Journal of Intelligent \& Robotic Systems},
  volume = {104},
  number = {2},
  pages = {22},
  doi = {10.1007/s10846-021-01527-7},
  urldate = {2026-08-31},
  langid = {english}
}

@article{jacewiczQuadrotorModelEnergy2022,
  title = {Quadrotor {{Model}} for {{Energy Consumption Analysis}}},
  author = {Jacewicz, Mariusz and {\.Z}ugaj, Marcin and G{\l}{\k e}bocki, Robert and Bibik, Przemys{\l}aw},
  year = 2022,
  month = sep,
  journal = {Energies},
  volume = {15},
  number = {19},
  pages = {7136},
  doi = {10.3390/en15197136},
  urldate = {2025-12-10},
  langid = {english}
}

@article{johnstonAdaptiveModifiedRISE2025a,
  title = {Adaptive {{Modified RISE Control}} for {{Quadrotors}}: {{Enhancing Trajectory Tracking Through Uncertainty Compensation}}},
  shorttitle = {Adaptive {{Modified RISE Control}} for {{Quadrotors}}},
  author = {Johnston, Kevin and Arrafi, Musabbir Ahmed and Kidambi, Krishna B. and Tiwari, Madhur},
  year = 2025,
  journal = {IEEE Access},
  volume = {13},
  pages = {169166--169177},
  doi = {10.1109/ACCESS.2025.3612215},
  urldate = {2026-08-31}
}

@article{kidambiRobustNonlinearControlBased2021,
  title = {Robust {{Nonlinear Control-Based Trajectory Tracking}} for {{Quadrotors Under Uncertainty}}},
  author = {Kidambi, Krishna Bhavithavya and Fermuller, Cornelia and Aloimonos, Yiannis and Xu, Huan},
  year = 2021,
  month = dec,
  journal = {IEEE Control Systems Letters},
  volume = {5},
  number = {6},
  pages = {2042--2047},
  doi = {10.1109/LCSYS.2020.3044833},
  urldate = {2026-09-04},
  copyright = {https://ieeexplore.ieee.org/Xplorehelp/downloads/license-information/IEEE.html},
  langid = {english}
}

@misc{kimMotionSpecificBatteryHealth2026,
  title = {Motion-{{Specific Battery Health Assessment}} for {{Quadrotors Using High-Fidelity Battery Models}}},
  author = {Kim, Joonhee and Park, Sanghyun and Kim, Donghyeong and Choi, Eunseon and Han, Soohee},
  year = 2026,
  month = mar,
  number = {arXiv:2603.12791},
  eprint = {2603.12791},
  primaryclass = {cs.RO},
  publisher = {arXiv},
  doi = {10.48550/arXiv.2603.12791},
  urldate = {2026-09-07},
  archiveprefix = {arXiv}
}

@article{lapandicMetaLearningAugmentedMPC2024,
  title = {Meta-{{Learning Augmented MPC}} for {{Disturbance-Aware Motion Planning}} and {{Control}} of {{Quadrotors}}},
  author = {Lapandi{\'c}, D{\v z}enan and Xie, Fengze and Verginis, Christos K. and Chung, Soon-Jo and Dimarogonas, Dimos V. and Wahlberg, Bo},
  year = 2024,
  journal = {IEEE Control Systems Letters},
  volume = {8},
  pages = {3045--3050},
  doi = {10.1109/LCSYS.2024.3520023},
  urldate = {2026-08-31}
}

@article{liuAdaptivePredefinedTimePrescribed2026,
  title = {Adaptive {{Predefined-Time Prescribed Performance Control With Application}} to a {{Quadrotor UAV}}},
  author = {Liu, Kang and Jiao, Lin and Zhang, Yu and Zhao, Pengyuan},
  year = 2026,
  journal = {IEEE Transactions on Aerospace and Electronic Systems},
  volume = {62},
  pages = {4751--4770},
  doi = {10.1109/TAES.2026.3654056},
  urldate = {2026-08-31}
}

@article{liuQuadrotorAggressiveControl2025,
  title = {Quadrotor {{Aggressive Control Based}} on {{Angular Acceleration Feedback}} and {{High-Fidelity Propulsion Model}}},
  author = {Liu, Hao and Lin, Defu and Ye, Jianchuan and Jiang, Tao and Huang, Jiangshuai},
  year = 2025,
  month = dec,
  journal = {IEEE Transactions on Industrial Electronics},
  volume = {72},
  number = {12},
  pages = {13584--13594},
  doi = {10.1109/TIE.2025.3572921},
  urldate = {2026-08-31}
}

@article{lopez-sanchezPIDControlQuadrotor2023,
  title = {{{PID}} Control of Quadrotor {{UAVs}}: {{A}} Survey},
  shorttitle = {{{PID}} Control of Quadrotor {{UAVs}}},
  author = {{Lopez-Sanchez}, Ivan and {Moreno-Valenzuela}, Javier},
  year = 2023,
  journal = {Annual Reviews in Control},
  volume = {56},
  pages = {100900},
  doi = {10.1016/j.arcontrol.2023.100900},
  urldate = {2026-08-31},
  langid = {english}
}

@article{martinsInnerouterFeedbackLinearization2022,
  title = {Inner-Outer Feedback Linearization for Quadrotor Control: Two-Step Design and Validation},
  shorttitle = {Inner-Outer Feedback Linearization for Quadrotor Control},
  author = {Martins, Lu{\'i}s and Cardeira, Carlos and Oliveira, Paulo},
  year = 2022,
  month = sep,
  journal = {Nonlinear Dynamics},
  volume = {110},
  number = {1},
  pages = {479--495},
  doi = {10.1007/s11071-022-07632-y},
  urldate = {2026-09-07},
  langid = {english}
}

@article{pattipatiOpenCircuitVoltage2014,
  title = {Open Circuit Voltage Characterization of Lithium-Ion Batteries},
  author = {Pattipati, B. and Balasingam, B. and Avvari, G.V. and Pattipati, K.R. and {Bar-Shalom}, Y.},
  year = 2014,
  month = dec,
  journal = {Journal of Power Sources},
  volume = {269},
  pages = {317--333},
  doi = {10.1016/j.jpowsour.2014.06.152},
  urldate = {2026-09-07},
  langid = {english}
}

@misc{renLearningAgileQuadrotor2026,
  title = {Learning {{Agile Quadrotor Flight}} in the {{Real World}}},
  author = {Ren, Yunfan and Zhu, Zhiyuan and Xing, Jiaxu and Scaramuzza, Davide},
  year = 2026,
  month = jul,
  number = {arXiv:2602.10111},
  eprint = {2602.10111},
  primaryclass = {cs.RO},
  publisher = {arXiv},
  doi = {10.48550/arXiv.2602.10111},
  urldate = {2026-08-31},
  archiveprefix = {arXiv}
}

@article{senguptaUrbanAirMobility2025,
  title = {Urban {{Air Mobility Research Challenges}} and {{Opportunities}}},
  author = {Sengupta, Raja and Bulusu, Vishwanath and Mballo, Chams Eddine and Onat, Emin Burak and Cao, Shangqing (Albert)},
  year = 2025,
  month = may,
  journal = {Annual Review of Control, Robotics, and Autonomous Systems},
  volume = {8},
  number = {1},
  pages = {407--431},
  doi = {10.1146/annurev-control-022823-031353},
  urldate = {2026-08-31},
  copyright = {http://creativecommons.org/licenses/by/4.0/},
  langid = {english}
}

@article{sunSafetyDrivenLocalizationUncertaintyDriven2024,
  title = {Safety-{{Driven}} and {{Localization Uncertainty-Driven Perception-Aware Trajectory Planning}} for {{Quadrotor Unmanned Aerial Vehicles}}},
  author = {Sun, Gang and Zhang, Xuetao and Liu, Yisha and Zhang, Xuebo and Zhuang, Yan},
  year = 2024,
  month = aug,
  journal = {IEEE Transactions on Intelligent Transportation Systems},
  volume = {25},
  number = {8},
  pages = {8837--8848},
  doi = {10.1109/TITS.2024.3361494},
  urldate = {2026-08-31}
}

@article{wengUnifiedOpencircuitvoltageModel2014,
  title = {A Unified Open-Circuit-Voltage Model of Lithium-Ion Batteries for State-of-Charge Estimation and State-of-Health Monitoring},
  author = {Weng, Caihao and Sun, Jing and Peng, Huei},
  year = 2014,
  month = jul,
  journal = {Journal of Power Sources},
  volume = {258},
  pages = {228--237},
  doi = {10.1016/j.jpowsour.2014.02.026},
  urldate = {2026-09-07},
  langid = {english}
}

@article{xingProbabilisticInferenceBasedEfficient2025,
  title = {A {{Probabilistic Inference-Based Efficient Path Planning Method}} for {{Quadrotors}}},
  author = {Xing, Siyuan and Xian, Bin and Jiang, Pengzhi},
  year = 2025,
  month = mar,
  journal = {IEEE Transactions on Industrial Electronics},
  volume = {72},
  number = {3},
  pages = {2810--2820},
  doi = {10.1109/TIE.2024.3440496},
  urldate = {2026-08-31}
}

@article{yacefOptimizationEnergyConsumption2017,
  title = {Optimization of {{Energy Consumption}} for {{Quadrotor UAV}}},
  author = {Yacef, F and Rizoug, N and Bouhali, O and Hamerlain, M},
  year = 2017,
  langid = {english}
}

@article{zhangEnergyConsumptionModels2021,
  title = {Energy Consumption Models for Delivery Drones: {{A}} Comparison and Assessment},
  shorttitle = {Energy Consumption Models for Delivery Drones},
  author = {Zhang, Juan and Campbell, James F. and Sweeney Ii, Donald C. and Hupman, Andrea C.},
  year = 2021,
  month = jan,
  journal = {Transportation Research Part D: Transport and Environment},
  volume = {90},
  pages = {102668},
  doi = {10.1016/j.trd.2020.102668},
  urldate = {2025-11-24},
  langid = {english}
}

@article{zhangSafetyPlanningControl2023,
  title = {A {{Safety Planning}} and {{Control Architecture Applied}} to a {{Quadrotor Autopilot}}},
  author = {Zhang, Wenyu and Jia, Jindou and Zhou, Sicheng and Guo, Kexin and Yu, Xiang and Zhang, Youmin},
  year = 2023,
  month = feb,
  journal = {IEEE Robotics and Automation Letters},
  volume = {8},
  number = {2},
  pages = {680--687},
  doi = {10.1109/LRA.2022.3230593},
  urldate = {2026-08-31}
}

@article{zhouEfficientRobustTimeOptimal2023,
  title = {Efficient and {{Robust Time-Optimal Trajectory Planning}} and {{Control}} for {{Agile Quadrotor Flight}}},
  author = {Zhou, Ziyu and Wang, Gang and Sun, Jian and Wang, Jikai and Chen, Jie},
  year = 2023,
  month = dec,
  journal = {IEEE Robotics and Automation Letters},
  volume = {8},
  number = {12},
  pages = {7913--7920},
  doi = {10.1109/LRA.2023.3322075},
  urldate = {2026-08-31}
}
\appendix
\begin{table}[h]
	\centering
	\caption{SOC-dependent two-RC equivalent-circuit parameters used in the
		numerical study. All quantities are pack-level values.}
	\label{tab:battery_lookup_parameters}
	
	\begin{tblr}{
			width = \linewidth,
			colspec = {
				X[0.55,c,m]
				X[0.90,c,m]
				X[0.85,c,m]
				X[0.85,c,m]
				X[0.85,c,m]
				X[0.6,c,m]
				X[0.6,c,m]
			},
			hlines = {1pt},
			vlines = {0.6pt},
		}
		SOC (\%) &
		$V_{\mathrm{oc}}$ (V) &
		$R_0$ ($\Omega$) &
		$R_1$ ($\Omega$) &
		$R_2$ ($\Omega$) &
		$\tau_1$ (s) &
		$\tau_2$ (s)\\
		
		5   & 12.035 & 0.040600 & 0.015795 & 0.009765 & 28 & 115\\
		15  & 12.658 & 0.035000 & 0.014274 & 0.008505 & 31 & 125\\
		30  & 13.378 & 0.031360 & 0.013104 & 0.007560 & 34 & 140\\
		50  & 14.300 & 0.028840 & 0.012168 & 0.006804 & 38 & 155\\
		70  & 15.220 & 0.027440 & 0.011466 & 0.006300 & 42 & 170\\
		85  & 15.910 & 0.028000 & 0.011700 & 0.006426 & 46 & 185\\
		100 & 16.600 & 0.030240 & 0.012636 & 0.007056 & 50 & 200\\
	\end{tblr}
\end{table}

The lookup-table SOC breakpoints are
\begin{equation}
	\mathbf{s}_b =
	\begin{bmatrix}
		0.05 & 0.15 & 0.30 & 0.50 & 0.70 & 0.85 & 1.00
	\end{bmatrix}.
	\label{eq:battery_soc_breakpoints}
\end{equation}

Piecewise-linear interpolation is used between consecutive breakpoints for
$V_{\mathrm{oc}}$, $R_0$, $R_1$, $R_2$, $\tau_1$, and $\tau_2$.
The corresponding polarization capacitances are obtained from
\begin{equation}
	C_j(s)=\frac{\tau_j(s)}{R_j(s)},
	\qquad j\in\{1,2\}.
	\label{eq:battery_lookup_capacitance}
\end{equation}
\end{document}